\documentclass[letterpaper]{article} 
\usepackage{aaai2026}  
\usepackage{times}  
\usepackage{helvet}  
\usepackage{courier}  
\usepackage[hyphens]{url}  
\usepackage{graphicx} 
\usepackage{natbib}  
\usepackage{caption} 
\usepackage{algorithm}
\usepackage{algorithmic}
\usepackage{enumitem}
\usepackage{multirow}
\usepackage{makecell}
\usepackage{amsmath}
\usepackage{booktabs}       
\usepackage{amsfonts}       
\usepackage{nicefrac}       
\usepackage{microtype}      
\newcommand{\cmark}{\checkmark}
\usepackage{etoolbox}
\usepackage{tocloft} 
\usepackage{placeins}  
\usepackage{newfloat}
\usepackage{listings}
\DeclareCaptionStyle{ruled}{labelfont=normalfont,labelsep=colon,strut=off} 
\floatstyle{ruled}
\newfloat{listing}{tb}{lst}{}
\floatname{listing}{Listing}
\usepackage[table]{xcolor}
\usepackage{hyperref}
\usepackage{fancyhdr}
\fancypagestyle{plain}{%
  \fancyhf{} 
  \fancyfoot[C]{\thepage} 

}
\title{CronusVLA: Towards Efficient and Robust Manipulation\\via Multi-Frame Vision-Language-Action Modeling}

\affiliations{
    \textbf{Hao Li}$^{1,2}$\thanks{Equal contribution, alphabetical order; Code: \href{https://github.com/InternRobotics/CronusVLA}{{InternRobotics/CronusVLA}}; Email: \texttt{lihaohn@mail.ustc.edu.cn}} \quad
    \textbf{Shuai Yang}$^{2,3}$\footnotemark[1] \quad
    \textbf{Yilun Chen}$^{2}$ \quad
    \textbf{Xinyi Chen}$^{2}$ \quad
    \textbf{Xiaoda Yang}$^{3}$ \quad
    \textbf{Yang Tian}$^{2}$ \\
    \textbf{Hanqing Wang}$^{2}$ \quad
    \textbf{Tai Wang}$^{2}$ \quad
    \textbf{Dahua Lin}$^{4}$ \quad
    \textbf{Feng Zhao}$^{1}$ \quad
    \textbf{Jiangmiao Pang}$^{2}$ \\
    $^1$University of Science and Technology of China \quad
    $^2$Shanghai Artificial Intelligence Laboratory \\
    $^3$Zhejiang University \quad
    $^4$The Chinese University of Hong Kong
}

\usepackage{bibentry}
\nocopyright

\begin{document}

\maketitle
\let\oldaddcontentsline\addcontentsline
\renewcommand{\addcontentsline}[3]{}
\begin{abstract}
Recent vision-language-action (VLA) models built on pretrained vision-language models (VLMs) have demonstrated strong performance in robotic manipulation. However, these models remain constrained by the single-frame image paradigm and fail to fully leverage the temporal information offered by multi-frame histories, as directly feeding multiple frames into VLM backbones incurs substantial computational overhead and inference latency. We propose \textbf{CronusVLA}, a unified framework that extends single-frame VLA models to the multi-frame paradigm. CronusVLA follows a two-stage process: \textbf{(1) Single-frame pretraining} on large-scale embodied datasets with autoregressive prediction of action tokens, establishing an effective embodied vision-language foundation; \textbf{(2) Multi-frame post-training}, which adapts the prediction of the vision-language backbone from discrete tokens to learnable features, and aggregates historical information via feature chunking. CronusVLA effectively addresses the existing challenges of multi-frame modeling while enhancing performance and observational robustness. To evaluate the robustness under temporal and spatial disturbances, we introduce \textbf{SimplerEnv-OR}, a novel benchmark featuring 24 types of observational disturbances and 120 severity levels. Experiments across three embodiments in simulated and real-world environments demonstrate that CronusVLA achieves leading performance and superior robustness, with a 70.9\% success rate on SimplerEnv, a 26.8\% improvement over OpenVLA on LIBERO, and the highest robustness score on SimplerEnv-OR. These results highlight the potential of efficient multi-frame adaptation in VLA models for more powerful and robust real-world deployment. \href{https://LiHaoHN.github.io/CronusVLA.github.io}{\textcolor{blue}{Project Website}}.

\end{abstract}

\begin{figure}[t]
    \centering
    \includegraphics[width=\linewidth]{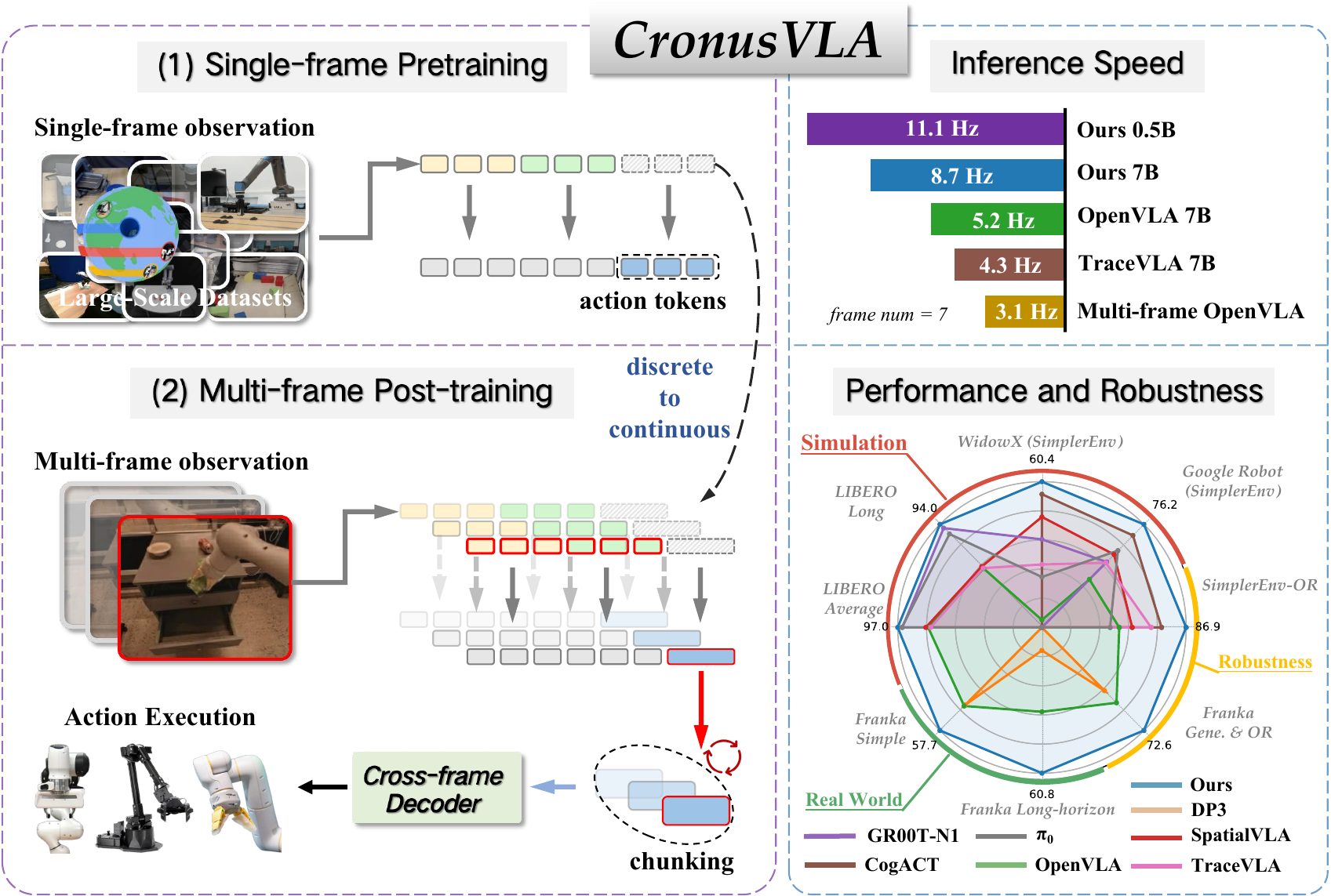}
    \caption{\textbf{CronusVLA} is a multi-frame modeling framework that includes single-frame pretraining on large-scale manipulation datasets and multi-frame post-training on cross-embodiment datasets. CronusVLA shows fast inference, high performance in simulation benchmarks and real-world experiments, and better observational robustness.}
    \label{fig:intro}
    \vspace{-11pt}
\end{figure}

\section{Introduction}
\label{sec:intro}
The rise of vision-language models (VLMs)~\cite{prismatic, qwen2vl} has paved the way for general vision-language-action (VLA) models by offering powerful backbones and pretrained vision-language representations. Recent VLA methods~\cite{openvla, spatialvla, Tracevla} primarily adapt advanced VLMs on large-scale heterogeneous manipulation datasets~\cite{oxe, droid} by re-engineering the tokenizer, while others~\cite{pi0, cogact} draw inspiration from low-level policy designs, incorporating techniques like specialized heads and action chunks for better performance.

Low-level policies~\cite{gr1,gr2,seer} have shown that leveraging multi-frame historical observations (i.e., RGB images captured from a single viewpoint over time) can considerably enhance performance. In fact, multi-frame information offers two notable advantages: (1) motion cues derived from consecutive observations \textit{help determine the current execution phase} and effectively resolve state ambiguities. (2) actions can be \textit{reliably inferred from prior consistent observations}, even when the current input is corrupted, demonstrating strong observational robustness during execution.

However, most existing VLA models~\cite{RT2, openvla, spatialvla}, built on the single-frame paradigm of vision-language models (VLMs), are typically trained using only a single current observation. Directly feeding multiple historical observation images into VLA models introduces two major challenges: (1) the self-attention computation in VLM backbones scales quadratically with the number of input tokens, making large-scale embodied pretraining computationally expensive; and (2) redundant visual tokens considerably degrade inference speed, limiting the feasibility of real-world deployment. RoboVLMs~\cite{robovlms} attempt to extend single-frame pretrained VLMs to the multi-frame paradigm by adopting memory-based LSTMs~\cite{LSTM} and training embodiment capabilities from scratch, following standard policy learning paradigms. However, they overlook the potential benefits of efficiently adapting single-frame pretrained models to the multi-frame paradigm, and lack quantitative evaluation of the performance gains brought by such multi-frame capabilities.

To facilitate efficient multi-frame modeling, we propose \textbf{CronusVLA}, a general framework for multi-frame training and inference, as illustrated in  Fig.\ref{fig:intro}. The approach comprises two key components: \textbf{(1) Single-frame Pretraining}: We first train a basic single-frame VLA model using standard autoregressive prediction over discrete action tokens, enabling more convenient utilization of large-scale heterogeneous embodied datasets and establishing an effective vision-language foundation. \textbf{(2) Multi-frame Post-training}: We introduce learnable features in the basic single-frame VLA model and conduct post-training on high-quality cross-embodiment datasets. This multi-frame post-training will replace multiple discrete action tokens with continuous learnable features and adapt the prediction of vision-language backbones from single-frame awareness to multi-frame. Extracting motion cues from multiple historical frames into a \textbf{feature chunking} enables effective temporal information aggregation, thereby improving efficiency. We further introduce the \textbf{feature modulator} and \textbf{multi-frame regularization}, which mitigate temporal imbalance and enhance convergence capability by reconstructing the influence of past frames within the model.

Current benchmarks, such as SimplerEnv~\cite{simplerenv} and LIBERO~\cite{Libero}, primarily focus on evaluating manipulation models across diverse tasks, objects, and scenes. They largely overlook the impact of observational disturbances on manipulation models, which is a critical issue for future real-world applications. We introduce the \textbf{SimplerEnv-OR} (Observational Robustness) benchmark, which enables quantitative evaluation of model robustness beyond standard training data augmentation, considering both temporal and spatial disturbances. It includes 24 types of observational disturbances with over 120 levels of severity, and evaluates performance across 2,300 trials.

In total, our approach offers three advantages: (1) Fast inference. CronusVLA predicts learnable features in a single forward pass without relying on autoregressive prediction, enabling faster action generation over previous VLA models. Cached historical learnable features further eliminate redundant computation during inference, resulting in substantial speed improvements. (2) High performance. Extensive experiments across three embodiments and diverse manipulation tasks in both the simulation and real world are conducted. CronusVLA achieves state-of-the-art performance on the simulation benchmark SimplerEnv with an average 70.9\% success rate, a 26.8\% overall improvement over OpenVLA on LIBERO. (3) Strong robustness. CronusVLA can maintain strong robustness against temporal and spatial disturbances. In our real-world experiments, it achieves a 72.6\% success rate under various challenges involving interference and occlusion. On our SimplerEnv-OR benchmark, the model consistently surpasses previous approaches in both robustness score and success rate.

Our contributions are summarized as follows:
\begin{itemize}[leftmargin=*]
\item We propose a general framework, CronusVLA, to extend VLA models to a multi-frame paradigm, which includes single-frame pertaining and multi-frame post-training.
\item We propose SimplerEnv-OR, a novel benchmark that enables quantitative evaluation of model robustness under observational disturbances.
\item Extensive experiments have demonstrated CronusVLA's leading performance and strong observational robustness across simulation and the real world.
\end{itemize}

\section{Related Works}
\label{sec:related}
\noindent\textbf{Vision-Language-Action models.}
Current VLA models usually integrate action generation into pretrained VLMs~\cite{prismatic,phi3}. By employing an action tokenizer, RT-2~\cite{RT2} and OpenVLA discretize 7D actions and generate them via autoregressive prediction. SpatialVLA~\cite{spatialvla} unifies the action space of various robots via adaptive action grids. 3D-VLA~\cite{3d_vla} and Magma~\cite{magma} unify action prediction and multimodal embodied tasks within a single model. These methods minimally adapt VLMs into a general manipulation policy by treating actions as discrete tokens. Instead, the remaining works~\cite{tinyvla, robovlms, instructvla} abandon the discrete formulation and do not aim to preserve the original paradigm of VLMs. They augment the original VLMs with additional action heads~\cite{LSTM, DiT}, meanwhile, train embodiment capabilities and continuous action prediction from scratch. Our method focuses on transferring a single-frame, discretely pretrained VLA model to the multi-frame paradigm with continuous action prediction.

\noindent\textbf{Multi-frame modeling for manipulation.} Most VLAs~\cite{openvla, palme, RT2} treat each action prediction as a temporally independent decision, and are also trained in a single-frame manner. While low-level policies~\cite{diffusionpolicy, ACT} incorporate multi-frame modeling by processing multiple images simultaneously based on their lightweight architectures. Other policies~\cite{gr1, seer} are pre-trained on large-scale video datasets~\cite{howto100m,ego4d} by interleaving multi-step action and multi-frame image prediction. However, directly applying this strategy to large-scale VLAs introduces considerable computational overhead. To address these limitations, Dita~\cite{hou2025dita} directly employs multiple RGB images. TraceVLA~\cite{Tracevla} uses visual prompting to draw past traces on the current observation. RoboFlamingo~\cite{roboflamingo} and RoboVLMs~\cite{robovlms} primarily adopt memory-based LSTMs and training embodiment capabilities from scratch to model temporal relations across frames. In contrast, our method builds on a single-frame pretrained VLA model and explicitly establishes multi-frame capabilities during post-training, which retains the single-frame perception meanwhile enabling multi-frame modeling.

\begin{figure*}[t]
    \centering
    \includegraphics[width=1\linewidth]{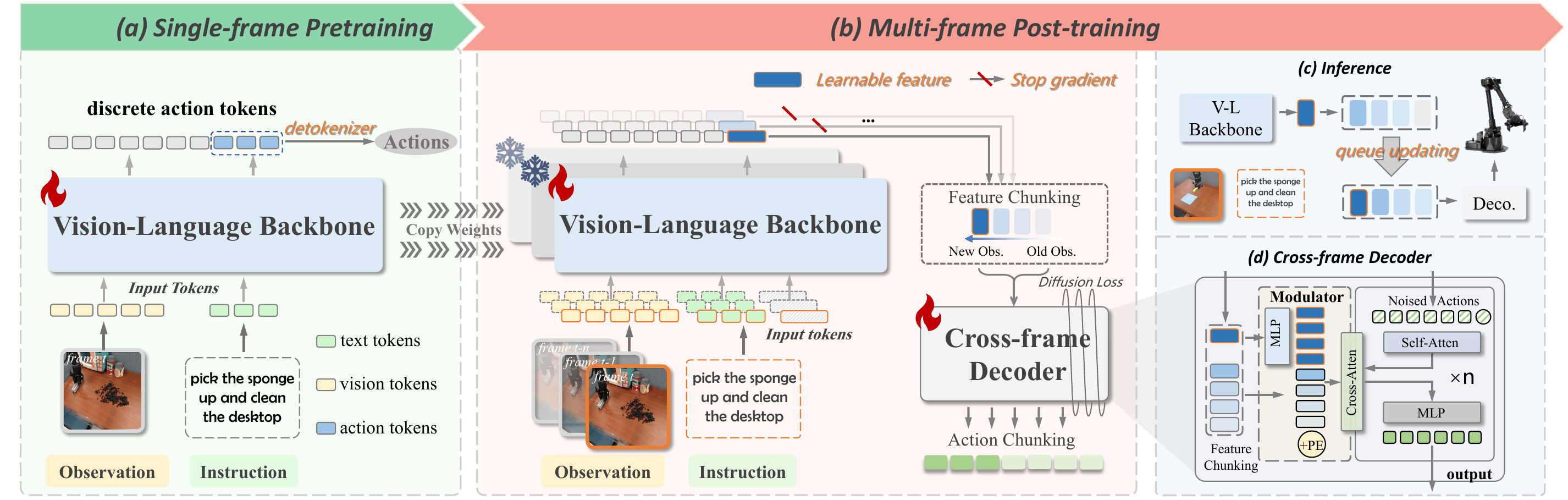}
    \caption{\textbf{Overview of CronusVLA framework.} (a) illustrates the single-frame pretraining of the basic single-frame VLA. By duplicating the model weights, we perform multi-frame post-training as shown in (b), where multi-frame modeling is achieved by aggregating learnable features from several preceding frames in a cross-frame decoder. In (c), a queue mechanism is conducted on feature chunking for fast inference. Details of the cross-frame decoder are illustrated in (d).}
    \label{fig:method}
    \vspace{-10pt}
\end{figure*}

\section{Methodology}
We present the details of our training strategy and model design in this section. In Sec.\ref{sec: Single-frame Pretraining}, we describe the single-frame training process. In Sec.\ref{sec: Multi-frame Encoding}, we detail the multi-frame post-training approach, including the vision-language backbone, the cross-frame decoder, and multi-frame regularization. Finally, Sec.\ref{sec: Action Adaptation} show our SimplerEnv-OR benchmark. The overview is illustrated in Fig.\ref{fig:method}.

\subsection{Single-frame Pretraining} \label{sec: Single-frame Pretraining}
As shown in Fig.\ref{fig:method} (a), our first step is to establish a basic single-frame VLA model. Off-the-shelf pretrained VLMs~\cite{prismatic,qwen2vl} are adapted by learning large-scale manipulation demonstrations $D_i = \{ I_t,a_t, l \})_{t=0}^{T_i}$, 
where $T_i$ is the length of episode $i$, $l$ is the language instruction, observation $I_t$ denotes a RGB image from a single camera at step $t$, and $a_t \in \mathbb{R}^{n}$ represents the corresponding actions. Following~\cite{openvla}, discrete action tokens are derived from continuous actions $a_{t}$ via the extended action tokenizer, which maps actions into 256 bins, and are trained using the next-token prediction objective. Given $I_t$ and $l$, the model predicts the next-step action tokens and detokenizes them, $a_{t} = \mathrm{VLA}(I_t, l)$. We observe that the single-frame pretraining effectively transfers the visual perception capabilities of vision encoders SigLip~\cite{siglip} and Dinov2~\cite{dinov2} to embodied scenes, which provides an effective vision-language foundation for multi-frame post-training. Meanwhile, single-frame pretraining can better preserve single-frame visual perception of VLMs and incur lower training costs on large-scale data, compared to directly pretraining in a multi-frame manner.

\subsection{Multi-frame Post-traing} \label{sec: Multi-frame Encoding}
\noindent\textbf{From discrete action tokens to feature chunking}. For our basic single-frame VLA, vision tokens $\left \{ v^i,i\in\left [  0,n_v\right ]  \right \}$ and text tokens $\left \{ l^i,i\in\left [  0,n_l\right ]  \right \}$ are causally computed in the vision-language backbone, which autoregressively predict discrete action tokens by summarizing information from all previous tokens. As shown in Fig.\ref{fig:method} (b), during multi-frame post-training, instead of generating discrete action tokens $a_t$, we introduce learnable features ${f}_t \in \mathbb{R}^d$ in the backbone’s hidden layers as continuous representations. This feature is designed to integrate the pretrained model’s embodied vision-language summarization capability and is computed as ${f}_t = \mathrm{VL}(I_t, l)$.
All images are still encoded by our vision-language backbone within a single-frame formulation, ensuring compatibility with standard VLM formulations. We introduce \textit{feature chunking} $ {F}_{t}^{M}=\{{f}_{t-M+1},\dots,{f}_{t-1},{f}_t\}=f_{t-M+1:t}$ to establish the association between different frames. It is a chunking of historical learnable features and can represent multi-frame images of $M$ steps at the feature-level. During training, we restructure $ M$-step image inputs at the batch level, enabling the vision-language backbone to independently process $B \times M$ single-frame inputs per iteration, where $B$ denotes the original batch size. During inference, as shown in Fig.\ref{fig:method} (c), we maintain the feature chunking using a first-in, first-out queue mechanism, which ensures fast inference by reusing prior vision-language computations.

\noindent\textbf{Cross-frame decoder.} The cross-frame decoder predicts action chunking by decoding the multi-frame information embedded in the feature chunking ${F}_{t}^{M}$, $a_{t:t+K-1}=\mathrm{Decoder}({F}_{t}^{M})$, as shown in Fig.\ref{fig:method} (d). We construct a DiT-based decoder composed of self-attention networks and MLP layers, and train it using a diffusion loss $\mathcal{L}$. To balance the contributions of the current and past learnable features, we employ a feature \textit{modulator} to dynamically modulate the learnable features. Specifically, the current feature $f_t\in \mathbb{R}^{d}$ is divided to match the number of past learnable features $f_{t-M+1:t-1}$ through channels splitting ($\mathrm{DIV}$), and then processed by modulator ($\mathrm{MD}$) to get the modulated feature ${Z}_{f}$:
\begin{equation}
    Z_f = \mathrm{MD}(F_t^M) = \mathrm{MLP}\left( {f}_{t-M+1:t-1},\tilde{f}_t \right)
\end{equation}
\begin{equation}
\label{eque_2}
    \tilde{f}_t = \mathrm{DIV}(f_t),\text{where } f_t\in \mathbb{R}^{d}, \tilde{f}_t\in \mathbb{R}^{(M-1) \times d},
\end{equation}
where $\mathrm{DIV}$ consists of a dimensionality-expanding Linear layer followed by a feature-splitting operation, $Z_f\in \mathbb{R}^{2\cdot (M-1) \times d'}$, $d'$ is the hidden dimension of the decoder. We further adopt a cross-attention mechanism to process noised actions and modulated features, which enables effective interaction. Specifically, $Z_f$ is fed into the cross-attention network and mapped to the keys and values, where noised actions $\hat{a}$ serve as queries. Noised actions are iteratively denoised conditioned on $Z_f$ for the final action output during inference.

\noindent\textbf{Post-training with multi-frame regularization.} We introduce \textit{the multi-frame regularization} to decouple the vision-language backbone from multi-frame modeling of the decoder, ensuring its training logic remains consistent with the single-frame paradigm. Specifically, the past learnable features ${f}_{t-M+1:t-1}$ of the feature chunking ${F}_{t}^{M}$ are treated as auxiliary inputs to the decoder, with their influence limited within the decoder, where $t$ is the execution step. Their gradient flows do not update the vision-language backbone, and serves solely as a regularization term to facilitate training:
\begin{equation}
\hat{f}_{t-M+1:t-1}
= \left\{ \mathrm{sg}\big(\mathrm{VL}(I_{t-k}, l)\big), k = 1, \dots, M{-}1 \right\},
\end{equation}
\begin{equation}
\mathcal{L} = 
\mathbb{E}_{\epsilon \sim \mathcal{N}(0, \mathbf{I}),i}
\left[
\left\|
\hat{\epsilon}^i - \epsilon_\theta\left(
t,\ \hat{f}_{t-M+1:t-1},\ f_t
\right)
\right\|_2
\right],
\end{equation}

where $\mathrm{sg}$ means the stop-gradient operation, $\mathrm{VL}$ is the vision-language backbone, $\mathcal{L}$ is the diffusion loss (multi-step denoising), $\hat{\epsilon}^i$ is the predicted $i$-step noise, associated with the noised actions $\hat{a}^i_{t:t+K-1}$. This method offers two advantages: (1) Extracting past learnable features without gradient computation reduces computational and memory overhead, enabling efficient training. (2) Updating on a single-frame basis preserves the pretrained single-frame perceptual capabilities and promotes faster convergence.

\subsection{SimplerEnv-OR Benchmark}
\label{sec: Action Adaptation}
\begin{figure}
    \centering
    \includegraphics[width=0.9\linewidth]{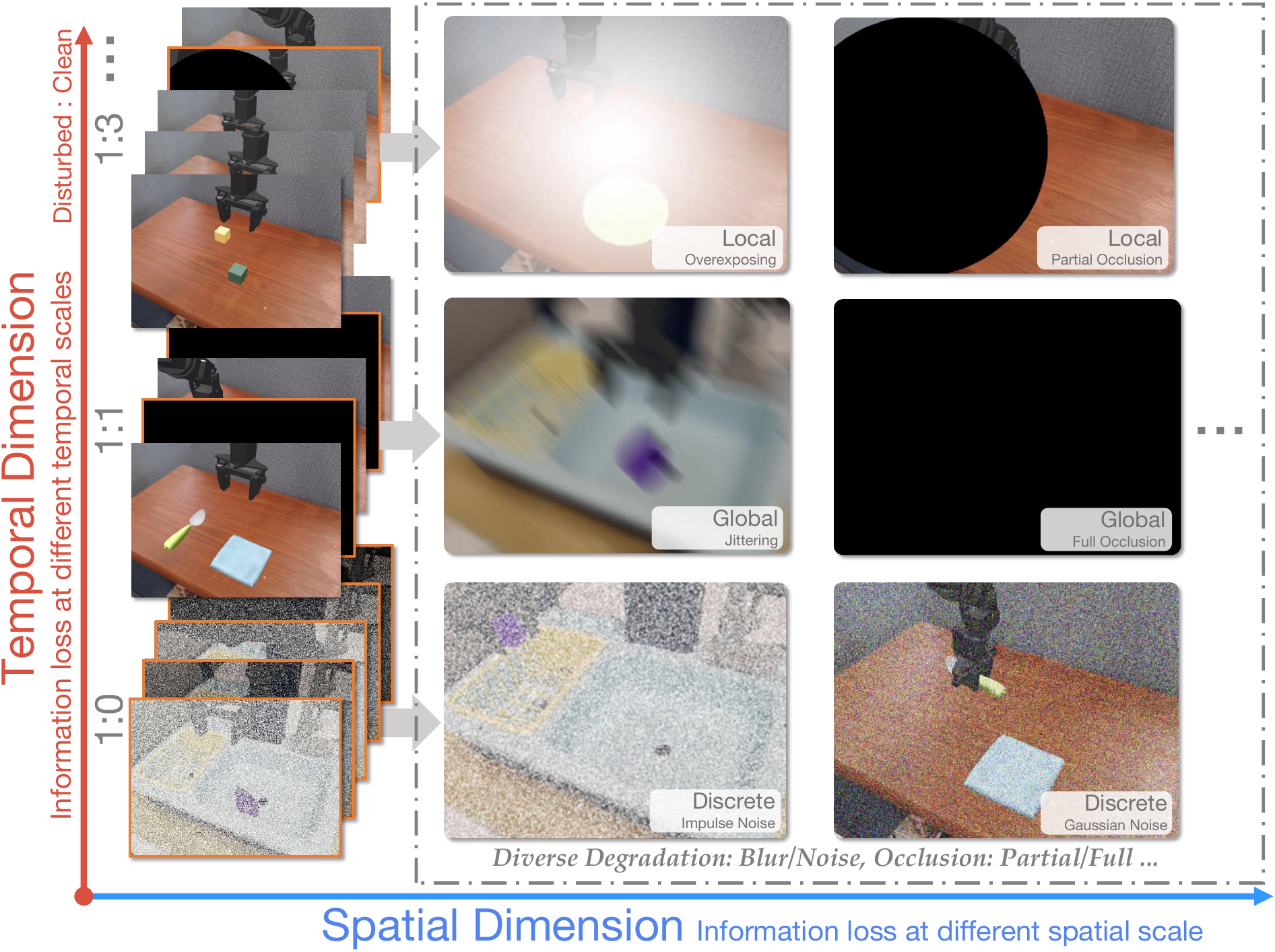}
    \caption{An illustration of the SimplerEnv-OR benchmark.}
    \label{fig:method2}
\end{figure}

As shown in Fig.\ref{fig:method2}, we introduce the SimplerEnv-OR benchmark. It is designed for quantitatively evaluating the observational robustness of VLAs by simulating different observational disturbances of the camera, which is beyond the simple visual augmentation strategies used in training. 

\noindent\textbf{Evaluation settings.}
SimplerEnv-OR extends the simulation environment of the WidowX Robot Visual Matching (WR-VM) setting in SimplerEnv~\cite{simplerenv}, and evaluates models trained on Bridge-v2~\cite{bridgev2}. We consider two disturbance dimensions: spatial and temporal. The spatial dimension introduces visual disturbances in different positions to simulate real-world camera artifacts, including Global (i.e., blurring, jittering, frame dropping, full occlusion), Local (i.e., overexposing, partial occlusions), and Discrete (i.e., noise, impulse). The temporal dimension assesses model robustness under varying disturbance frequencies, including Constant (1:0), Cyclic (1:1), and Sparse (1:3, 1:5, ...). Under all disturbance settings, VLAs must complete the same tasks as WR-VM. Details are provided in Appendix.\ref{sec_details_benchmark}.

\noindent\textbf{Robustness score.}
To quantify robustness, we define a robustness score $R\text{-}Score$ based on relative performance maintenance under various disturbances. Let $SR$ denote the average success rate of the original WR-VM tasks, and $SR^i$ denote the success rate under the disturbance setting $i$. The robustness score is computed as: $R\text{-}Score^{i} = 100 * \frac{SR^i}{SR}$. Noting that each setting includes 200 to 400 trials, ensuring a stable and reliable evaluation.

\section{Experiment}
\subsection{Results in Simulation}
\label{sec:main_results_for_zero_shot}

\renewcommand{\arraystretch}{1.1}
\begin{table*}[t]
 \centering
 \resizebox{\textwidth}{!}{
\begin{tabular}{lcccccccccccccccc}
\toprule
\multirow{5}{*}{Methods} & \multicolumn{10}{c}{Google Robot}  & \multicolumn{5}{c}{WidowX Robot}  & \multirow{5}{*}{\textbf{Avg}} \\ \cmidrule(lr){2-11}\cmidrule(lr){12-16}
& \multicolumn{2}{c}{\begin{tabular}[c]{@{}c@{}}Open/Close \\ Drawer\end{tabular}} & \multicolumn{2}{c}{\begin{tabular}[c]{@{}c@{}}Put in \\ Drawer\end{tabular}} & \multicolumn{2}{c}{\begin{tabular}[c]{@{}c@{}}Pick \\Coke Can\end{tabular}} & \multicolumn{2}{c}{\begin{tabular}[c]{@{}c@{}}Move\\ Near\end{tabular}} & \multicolumn{2}{c}{\textbf{Avg}}  & \begin{tabular}[c]{@{}c@{}}Put \\ Spoon\end{tabular} & \begin{tabular}[c]{@{}c@{}}Put \\ Carrot\end{tabular} & \begin{tabular}[c]{@{}c@{}}Stack \\ Blocks\end{tabular} & \begin{tabular}[c]{@{}c@{}}Put\\ Eggplant\end{tabular} & \textbf{Avg} & \\ \cmidrule(lr){2-11}\cmidrule(lr){12-16}
& VM & VA & VM  & VA & VM & VA & VM  & \multicolumn{1}{c}{VA} & VM & VA & \multicolumn{5}{c}{VM}  &   \\ \cmidrule(l){1-1}\cmidrule(lr){2-11}\cmidrule(lr){12-16}\cmidrule(lr){17-17}
RT-1-X$^{*}$~\cite{oxe}  & 59.7 & 29.4 & 21.3 & 10.1 & 56.7 & 49.0 & 31.7  & \multicolumn{1}{c}{32.3} & 42.4 & 30.2 & 0.0 & 4.2 & 0.0  & \multicolumn{1}{c}{0.0} & 1.1 & 24.6 \\
RT-2-X$^{*}$~\cite{oxe} & 25.0 & 35.5& 3.7 & 20.6  & 78.7 & 82.3& 77.9  & \multicolumn{1}{c}{79.2}   & 46.3 & 54.4 & - & -  & - & \multicolumn{1}{c}{-}& - & -\\
Octo-Base$^{*}$~\cite{octo}   & 22.7 & 1.1 & 0.0 & 0.0   & 17.0 & 0.6 & 4.2   & \multicolumn{1}{c}{3.1} & 11.0 & 1.2  & 15.8 & 12.5  & 0.0  & \multicolumn{1}{c}{41.7}   & 17.5 & 9.9 \\ \cmidrule(l){1-1}\cmidrule(lr){2-11}\cmidrule(lr){12-16}\cmidrule(lr){17-17}

RoboVLMs (2B)~\cite{robovlms}  & 44.9 & 10.3& 27.8& 0.0  & 76.3 & 50.7& 79.0  & \multicolumn{1}{c}{62.5}   & 57.0 & 30.9 & 50 & 37.5& 0.0  & \multicolumn{1}{c}{83.3} & 42.7  & 43.5   \\

SpatialVLA (3B)$^{*}$~\cite{spatialvla}  & \textbf{54.6} & \textbf{39.2}& 0.0 & 6.3  & 79.3 & 78.7 & 90.0  & \multicolumn{1}{c}{\textbf{83.0}}   & 56.0 & 51.8 & 20.8 & 37.5& 41.7  & \multicolumn{1}{c}{\textbf{83.3}} & 45.8  & 51.2   \\

$\pi_0$ (3B)~\cite{pi0} & 38.3 & 25.6& - & -   & 72.7 & 75.2& 65.3  & \multicolumn{1}{c}{63.7}   & - & - & 25.0  & 16.7  & 12.5 & \multicolumn{1}{c}{29.2}   & 20.9 & -  \\

$\pi_0$-FAST (3B)~\cite{pertsch2025fast} & 42.9 & 31.3& - & -   & 75.3 & 77.6& 67.5  & \multicolumn{1}{c}{68.2}   & - & - & 29.1  & 21.9  & 10.8 & \multicolumn{1}{c}{66.6}   & 32.1 & -  \\

GR00T-N1.5 (2B)~\cite{bjorck2025gr00t} & 27.8 & 35.8& 7.4 & 4.0   & 51.7 & 69.3& 54.0  & \multicolumn{1}{c}{68.7}   &  35.2 & 44.5 & \textbf{75.3}  & \textbf{54.3}  & \textbf{57.0} & \multicolumn{1}{c}{61.3}   & \textbf{61.9}
 & 47.2  \\\rowcolor{gray!15}
\textbf{Ours (0.5B)$^{*}$}   & 50.5 & 36.8   & \textbf{42.6} & \textbf{21.7} & \textbf{96.0} & \textbf{94.6}   & \textbf{93.0} & \multicolumn{1}{c}{78.0}  & \textbf{70.5} & \textbf{57.8} & 45.8& 33.3  & 0.0  & \multicolumn{1}{c}{79.2}   & 39.6 & \textbf{56.0}   \\ \cmidrule(l){1-1}\cmidrule(lr){2-11}\cmidrule(lr){12-16}\cmidrule(lr){17-17}

OpenVLA (7B)$^{*}$~\cite{openvla}  & 59.7 & 23.5 & 0.0 & 2.9 & 25.7 & 54.1 & 55.0 & \multicolumn{1}{c}{63.0}   & 35.1 & 35.9 & 8.3 & 4.2  & 0.0 & \multicolumn{1}{c}{0.0}& 3.1 & 24.7\\
CogACT (7B)$^{*}$~\cite{cogact}& 71.8 & 28.3 & 50.9 & 46.6  & 91.3 & 89.6   & \textbf{85.0} & \multicolumn{1}{c}{\textbf{80.8}}   & 74.8 & 61.3 & \textbf{75.0}& 50.0  & 16.7   & \multicolumn{1}{c}{79.2}   & 55.2 & 63.8   \\
TraceVLA (7B)$^{*}$~\cite{Tracevla} & 63.1 & \textbf{61.6}   & 11.1   & 12.5  & 45.0 & 64.3& 63.8  & \multicolumn{1}{c}{60.6}   & 45.8 & 49.8 & 12.5 & 16.6  & 16.6 & \multicolumn{1}{c}{65.0}   & 27.7 & 41.1   \\
OpenVLA-OFT (7B)~\cite{openvla-oft} & 47.2 & 12.2   & 0.9   & 0.5  & 72.3 & 65.3& 69.6  & \multicolumn{1}{c}{59.0}   & 47.5 & 34.3 & 12.5 & 4.2  & 4.2 & \multicolumn{1}{c}{100.0}   & 30.2 & 37.3   \\
Magma (8B)$^{*}$~\cite{magma}   & 58.9 & 59.0 & 8.3 & 24.0 & 75.0 & 68.6 & 53.0 & \multicolumn{1}{c}{78.5}   & 48.8 & 57.5 & 37.5 & 29.2  & 20.8 & \multicolumn{1}{c}{91.7}& 44.8 & 50.4\\\rowcolor{gray!15}
\textbf{Ours (7B)$^{*}$}  & \textbf{77.8} & 58.7& \textbf{64.8} & \textbf{65.1} & \textbf{95.7} & \textbf{94.2} & 76.0 & \multicolumn{1}{c}{77.0}  & \textbf{78.6} & \textbf{73.8} & 66.7 & \textbf{54.2} & \textbf{20.8} & \multicolumn{1}{c}{\textbf{100.0}}   & \textbf{60.4} & \textbf{70.9}\\ \bottomrule
\end{tabular}
 }
\caption{\textbf{Performance comparison on SimplerEnv.} Experiments are conducted across 12 tasks. Sizes of the original LLMs are listed, with \% omitted. $^{*}$ indicates evaluating all 12 tasks using a co-trained checkpoint; otherwise, two checkpoints are used to evaluate the Google Robot and WidowX setups separately.}

\label{tab:main_results}
 \vspace{-10pt}
\end{table*}

\noindent\textbf{Implementation details.}
In this section, we primarily investigate the performance of our post-trained model. After pretraining the basic single-frame VLA following \cite{openvla, belkhale2024minivla} with the OXE dataset~\cite{oxe}, we select two high-quality datasets, Bridge-v2~\cite{bridgev2} and Fractal~\cite{RT1} datasets, to conduct multi-frame post-training, which include about 148k episodes and 5M multi-frame clips. Our CronusVLA 7B is built on 7B Llama 2~\cite{llama2}, and CronusVLA 0.5B is built on Qwen2.5 0.5B~\cite{yang2024qwen2}, they both employ Dinov2 and SigLip as vision encoders. CronusVLA is built on a third-person camera and a text instruction. CronusVLA 7B is configured with a default of 6 past frames, while CronusVLA 0.5B utilizes 3 past frames. All experiments are based on A100 GPUs. More training details are included in Appendix.

\noindent\textbf{Evaluation setup in SimplerEnv.} We conduct experiments within SimplerEnv \cite{simplerenv}, a benchmark to evaluate models across various tasks with the WidowX Robot (WR) and the Google Robot (GR). The GR environment includes two settings, Visual Matching (VM) and Variant Aggregation (VA). WidowX Robot environment only includes the VM setting. SimplerEnv covers more than 2k trails across different scenarios, objects, and tasks. We report the average success rate of each task. RT-1-X~\cite{oxe, RT1}, RT-2-X~\cite{oxe, RT2}, and Octo-Based~\cite{octo} are early baselines, OpenVLA~\cite{openvla}, CogACT~\cite{cogact}, and Magma~\cite{magma} are trained on subsets of the OXE. $\pi_0$, $\pi_0$-FAST and GR00T-N1.5 are VLAs with additional robot state input. RoboVLMs~\cite{robovlms}, SpatialVLA~\cite{spatialvla}, and TraceVLA~\cite{Tracevla} are all trained on the Fractal and Bridge-V2 datasets.

\noindent\textbf{Experimental results on SimplerEnv.} Main results are illustrated in Tab.\ref{tab:main_results}. For Google Robot setting, our CronusVLA-7B achieves the highest average success rate, with 78.6 in the VM and 73.8 in the VA, surpassing TraceVLA and RoboVLMs (also multi-frame VLAs), by +71.6\% and +37.9\% relative VM score, +48.2\% and +138.8\% relative VA scores. Notably, our model achieves strong performance on not only simple tasks \textit{Pick Coke Can} and \textit{Open/Close Drawer}, but also the long-horizon task \textit{Put in Drawer}, which requires sequential actions of opening the drawer and placing the apple in it. Most previous approaches have failed to attain high success rates in \textit{Put in Drawer}, while our method effectively improves its VM success rate to 64.8 and the VA to 65.1. For the WidowX, CronusVLA 7B also achieves a high average success rate, +41.5\% higher than SpatialVLA and +9.4\% higher than CogACT. CronusVLA 0.5B, with a smaller language backbone, outperforms many prior models trained on larger (from 2B to 7B) language models. It achieves the best results in \textit{Pick Coke Can} and \textit{Move Near} over all other models, suggesting that excessive parameters may not always be beneficial, emphasizing the value of effective modeling.

\begin{table}[t]
  \centering
  \resizebox{\linewidth}{!}{
    \begin{tabular}{c|cccc|c}
      \toprule
      Methods & Spatial & Object & Goal & Long & Ave. \\ \midrule
      Diffusion Policy~\cite{diffusionpolicy} & 78.3 & 92.5 & 68.3 & 50.5 & 72.4 \\
      Dita~\cite{hou2025dita} & 84.2 & 96.3 & 85.4 & 63.8 & 82.4 \\
      OpenVLA~\cite{openvla} & 84.7 & 88.4 & 79.2 & 53.7 & 76.5 \\
      TraceVLA~\cite{Tracevla} & 84.6 & 85.2 & 75.1 & 54.1 & 74.8 \\
      SpatialVLA~\cite{spatialvla} & 88.2 & 89.9 & 78.6 & 55.5 & 78.1 \\ \midrule
      $\pi_0$-FAST~\cite{pertsch2025fast}&96.4 & 96.8 & 88.6 & 60.2 & 85.5\\ 
      GR00T-N1~\cite{bjorck2025gr00t} &94.4 & 97.6 & 93.0 & \underline{90.6} & 93.9\\ 
      $\pi_0$~\cite{pi0} &96.8 & \underline{98.8} & 95.8 & 85.2 & 94.2\\ 
      $\pi_{0.5} +KI$~\cite{intelligence2025pi_} &\textbf{98.0} & 97.8 & \underline{95.6} & 85.8 & 94.3\\ \midrule
      \rowcolor{gray!15}
      \textbf{Ours(7B)} & \underline{97.3} & \textbf{99.6} & \textbf{96.9} & \textbf{94.0} & \textbf{97.0} \\
      \bottomrule
    \end{tabular}
  }
    \caption{\textbf{Main results in LIBERO~\cite{Libero}}, the average success rates are reported. More details are in Tab.\ref{tab:detailed_libero_results}.}
    \label{tab:libero}
  \vspace{-5pt}
\end{table}

\begin{table}[t]
\centering
\renewcommand{\arraystretch}{1.2}
\resizebox{\linewidth}{!}{
\begin{tabular}{lcccccccc}
\toprule
& \multicolumn{6}{c}{Temporal Dimension} & \multicolumn{2}{c}{{\textit{WidowX Robot}}} \\
& \multicolumn{2}{c}{Constant (1:0)} & \multicolumn{2}{c}{Cyclic (1:1)} & \multicolumn{2}{c}{Sparse (1:3)} & \multicolumn{2}{c}{{\textit{VM}}} \\
\cmidrule(lr){2-7}
Methods & R-Score & SR & R-Score & SR & R-Score & SR & - & \textit{SR} \\
\midrule
$\pi_0$~\cite{pi0}  & 43.5 & 9.1 & 36.8 & 7.7 & 34.9 & 7.3  & - & \textit{20.9} \\
TraceVLA~\cite{Tracevla}  & 59.2 & 16.4 & 62.5 & 17.3 & 78.0 & 21.6  & - & \textit{27.7} \\
RoboVLMs~\cite{robovlms}     & 47.6 & 20.3 & 57.3 & 24.5 & 78.7 & 33.6  & - & \textit{42.7} \\
SpatialVLA~\cite{spatialvla}   & 44.9 & 20.6 & 48.0 & 22.0 & 63.1 & 28.9 & - & \textit{45.8} \\
CogACT~\cite{cogact}       & 53.3 & 29.4 & 66.1 & 36.5 & 80.2 & 44.3 & - & \textit{55.2} \\\midrule
\rowcolor{gray!15}
\textbf{Ours (7B)} & \textbf{61.2} & \textbf{37.0} & \textbf{86.7} & \textbf{52.3} & \textbf{96.2} & \textbf{58.1} & - & \textit{60.4} \\
\midrule
\midrule
& \multicolumn{6}{c}{Spatial Dimension} & \multicolumn{2}{c}{\multirow{2}{*}{Total Avg.}} \\
& \multicolumn{2}{c}{Global} & \multicolumn{2}{c}{Local} & \multicolumn{2}{c}{Discrete} & & \\
\cmidrule(lr){2-7}
Methods & R-Score & SR & R-Score & SR & R-Score & SR & \textbf{R-Score} & SR \\
\midrule
$\pi_0$ ~\cite{pi0} & 42.6 & 8.9 & 28.2 & 5.9 & 52.1 & 10.9  & 41.1 & 8.6 \\
TraceVLA~\cite{Tracevla}  & 58.1 & 16.1 & 65.3 & 18.1 & 81.9 & 22.7  & 65.8 & 18.2 \\
RoboVLMs~\cite{robovlms}     & 54.7 & 23.4 & 83.3 & 35.6 & 76.8 & 32.8 & 67.4 & 28.8 \\
SpatialVLA~\cite{spatialvla}   & 57.6 & 26.4 & 50.0 & 22.9 & 52.4 & 24.0 & 54.4 & 24.9 \\
CogACT~\cite{cogact}      & 60.2 & 33.2 & 80.5 & 44.4 & \textbf{87.4} & 48.3 & 72.1 & 39.8 \\\midrule
\rowcolor{gray!15}
\textbf{Ours (7B)} & \textbf{85.4} & \textbf{51.6} & \textbf{96.6} & \textbf{58.3} & 80.2 & \textbf{48.4} & \textbf{86.9} & \textbf{52.4} \\
\bottomrule
\end{tabular}
}
\caption{\textbf{Observational robustness test on SimplerEnv-OR.} Top: Temporal dimension with different frequencies. Bottom: Spatial dimension with different patterns. R-Score is the pre-defined score of robustness. SR denotes the success rate (\%).}
\label{tab:wrvm_cleaned}
  \vspace{-10pt}
\end{table}

\begin{figure*}[t]
    \centering
    \includegraphics[width=0.95\linewidth]{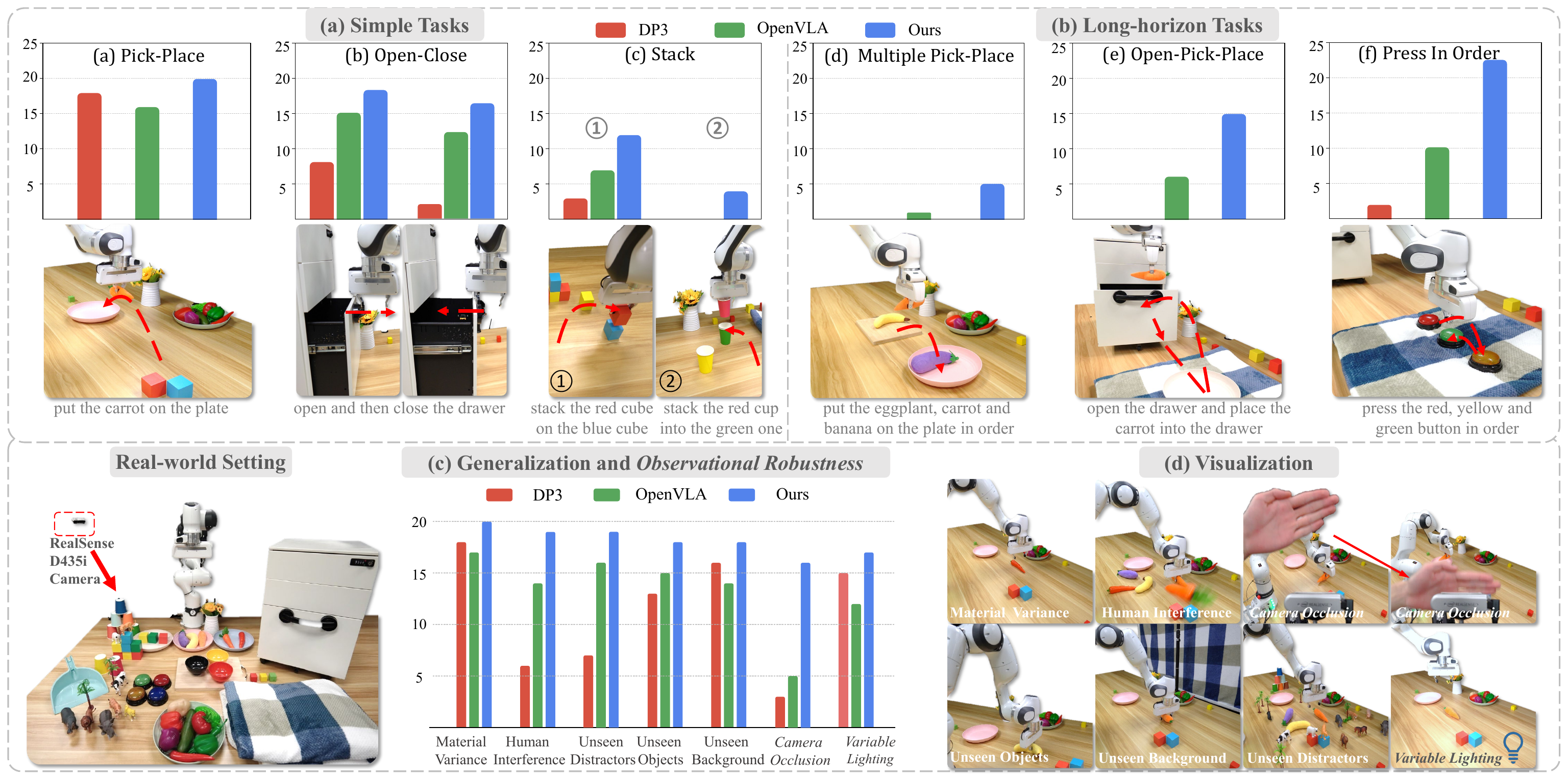}
    \caption{\textbf{Real-world experiment.} Evaluation of basic pick-and-place capabilities is in (a), long-horizon tasks of (b) demonstrate the advantages of multi-frame modeling in handling temporally dependent manipulations, and (c) generalization and robustness tests, particularly under camera occlusion and various disturbances, highlight the robustness of our model.}
    \label{fig:real_world_task}
    \vspace{-5pt}
\end{figure*}

\noindent\textbf{Evaluation setup and experimental results in LIBERO.} We evaluate CronusVLA on the LIBERO~\cite{Libero} benchmark. LIBERO comprises four task suites, including Spatial, Object, Goal, and Long. Based on the post-trained weights, CronusVLA is fully finetuned on each suite. We compare it with Diffusion Policy~\cite{diffusionpolicy}, Dita~\cite{hou2025dita}, OpenVLA, TraceVLA, and SpatialVLA, as well as other models conditioned on an additional wrist-view image and the robot state, including $\pi_0$, $\pi_0$-FAST, GR00T-N1, and $\pi_{0.5}$. As shown in Tab.\ref{tab:libero}, our model with only an extra wrist-view input achieves the highest average success rate of 97.0\%, attaining good performance across almost all individual suites and a remarkable 94.0\% on Long (a +40.3\% over OpenVLA), which demonstrates our robust in-domain learning and long-horizon execution capabilities.

\noindent\textbf{Robustness testing on SimplerEnv-OR.} We evaluate the observational robustness of several models: RoboVLMs (also a multi-frame VLA), SpatialVLA (spatial modeling VLA), CogACT (prior SOTA model), TraceVLA (designed for temporal modeling), $\pi_0$ (popular baseline). All results are summarized in Tab.\ref{tab:wrvm_cleaned}. In the \textit{Temporal Dimension,} as the disturbance frequency decreases from Constant (1:0) to Sparse (1:3), all models exhibit improved R-Scores. Single-frame models ($\pi_0$, SpatialVLA and CogACT) tend to produce out-of-distribution actions under high-frequency disturbances, leading to task failure. RoboVLMs and TraceVLA, despite being multi-frame, heavily rely on accurate historical information and often default to inaction or repeated probing when disturbed. In contrast, CronusVLA demonstrates superior robustness due to effective multi-frame modeling, achieving strong resistance under Constant (1:0) and near immunity under Sparse (1:3) with 96.2 R-Score. In the \textit{Spatial Dimension}, SpatialVLA is highly sensitive to local disturbances due to its reliance on zero-shot depth estimation, while CogACT and RoboVLMs are more affected by global disturbances. CronusVLA consistently outperforms others under these two types, exhibiting strong robustness. Notably, although SpatialVLA outperforms RoboVLMs on the original SimplerEnv benchmark, it underperforms on SimplerEnv-OR, with lower SR and R-Score, which highlights RoboVLMs’ robustness advantage as a multi-frame model. Overall, SimplerEnv-OR provides quantitative evidence of CronusVLA’s and other VLAs' robustness under observational disturbances.

\subsection{Real-world Experimental Results}

\noindent\textbf{Evaluation setup within Franka platform.} As shown in Fig.\ref{fig:real_world_task}, we evaluate our method on several real-world tasks with the Franka Research 3 Robot. We utilize a third-person camera for visual input. Three task suites are designed: \textit{(1) Simple Tasks}, involves the picking and placing objects, opening and closing the drawer, and stacking cubes or cups; \textit{(2) Long-horizon Tasks}, requires coordinated multi-step manipulation and includes putting multiple objects, placing objects into a drawer and pressing buttons in a specific order; and \textit{(3) Generalization and Observational Robustness} tasks, evaluating performance on unseen objects, novel instructions, camera occlusions, distractor objects and so on. We manually collect 50 episodes for each task, and successful rollouts among 25 trials are reported. We implement DP3~\cite{dp3}, and OpenVLA~\cite{openvla} finetuned on these demonstrations, our CronusVLA 7B is finetuned from the post-trained weight. More details are in Appendix.

\noindent\textbf{Experimental results within Franka platform.} CronusVLA outperforms other models across almost all tasks. For \textit{Simple Tasks}, all three policies perform well when handling simple objects; however, for tasks requiring more precise manipulation, such as stacking one block (or cup) on another, CronusVLA shows better in-domain performance. For \textit{Long-horizon Tasks}, CronusVLA exhibits stronger long-horizon learning capabilities, achieving consistently better performance than DP3 and OpenVLA under limited expert demonstrations. Notably, in \textit{press buttons in order}, OpenVLA tends to press the same button multiple times, indicating state confusion in long-horizon tasks due to the absence of multi-frame information, while CronusVLA has inherent temporal awareness for button pressing. \textit{Generalization and Observational Robustness} tasks are illustrated in Fig.\ref{fig:real_world_task} (c) and (d). In all distracted situations, our model shows the best performance. DP3 is sensitive to distractions and human interference. In the \textit{Camera Occlusions}, DP3 and OpenVLA are adversely affected by frequent visual dropouts due to their reliance on precise per-step observations, while our multi-frame modeling effectively withstands such disturbances.

\subsection{Ablations Study}
\noindent\textbf{Ablations on post-training strategies}. As shown in Tab.~\ref{table:training_strategy}, we evaluate the effectiveness of different multi-frame post-training strategies (with total 7 frames) by comparing the following configurations: (1) \textit{Baseline}: Our basic single-frame VLA model, further post-trained without multi-frame modeling. (2) \textit{+M.F.}: Directly feeding multiple images into the vision-language backbone and post-training in a discrete action space. This yields a margin performance gain (only +4.5\%) but considerably reduces inference speed (-40.3\%). (3) \textit{+M.F. +Dec.}: Replacing discrete action prediction with continuous feature prediction, training only the decoder during multi-frame modeling. This setting achieved substantial performance improvement and considerably boosts inference speed (8.73 Hz vs. 3.09 Hz), thanks to the elimination of autoregressive decoding and the caching of feature chunking. (4) \textit{+M.F. +Dec. +V.L.}: Enabling training of both the decoder and vision-language backbone under multi-frame post-training also improves the overall performance, attributed to the enhanced fitting ability. (5) \textit{+M.F. +Dec. +V.L. +Reg. (Ours)}: Utilizing the multi-frame regularization can effectively leverage single-frame perception of vision-language backbone while enabling sufficient multi-frame modeling of the decoder, leading to superior performance. Additionally, we observe that multi-frame regularization considerably improves the convergence (details in Appendix).

\begin{table}[t]
  \centering
  \resizebox{\linewidth}{!}{
    \begin{tabular}{l|ccccc}
      \toprule
       Settings & G-VM & G-VA & W-VM &Ave. & Speed (Hz) \\ 
      \midrule
      Baseline & 43.3 & 39.4& 10.4& 31.0 & 5.18 \\
      +M.F. & 45.5 &  47.4& 4.2& 32.4 & 3.09 \\
      +M.F. +Dec. & 52.2 & 58.1& 34.4& 48.2 & 8.73 \\ 
      +M.F. +Dec. +V.L. & 72.7 & 72.6& 56.3& 67.2 & 8.73 \\ 
      +M.F. +Dec. +V.L. +Reg. & \textbf{78.6} & \textbf{73.8}& \textbf{60.4}& \textbf{70.9} & \textbf{8.73} \\
      \bottomrule
    \end{tabular}
  }
  \caption{\textbf{Ablation on post-training strategies.} Success rates on SimplerEnv and inference speeds are shown. We discuss adding multiple images, training decoder, training vision-language backbone, and multi-frame regularization.}
\label{table:training_strategy}
\end{table}

\begin{table}[t]
  \centering
  \resizebox{0.9\linewidth}{!}{
    \begin{tabular}{c|l|cccc}
      \toprule
      \# & Settings & G-VM & G-VA & W-VM & Ave. \\
      \midrule
      1 & w/o. cross-atten & 76.4 & 71.2 & 57.3 & 68.3 \\
      2 & w/o. modulator & 68.7 & 61.4 & 60.4 & 63.5 \\\cmidrule(l){1-1}\cmidrule(lr){2-2}\cmidrule(lr){3-6}
      3 & MLP-based decoder & 66.4 & 53.6 & 35.4 & 51.8 \\ 
      4 & SiT-based decoder & 75.0 & 69.5 & 58.4 & 67.6 \\\cmidrule(l){1-1}\cmidrule(lr){2-2}\cmidrule(lr){3-6}
      5 & \textbf{Ours} & \textbf{78.6} & \textbf{73.8} & \textbf{60.4} & \textbf{70.9} \\
      \bottomrule
    \end{tabular}
  }
  \caption{\textbf{Ablation on cross-frame decoder designs.} Based on CronusVLA 7B with a past frame number of 6.}
  \label{table:decoder}
  \vspace{-10pt}
\end{table}

\begin{figure}[t]
    \centering
    \includegraphics[width=0.95\linewidth]{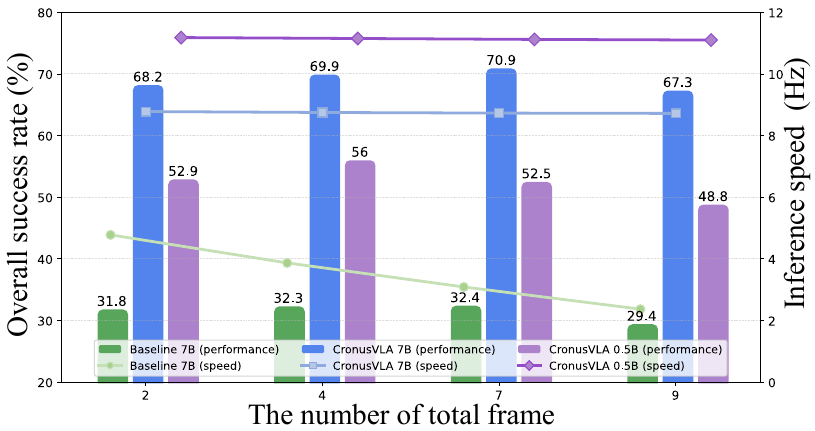}
    \caption{\textbf{Impact of frame number.} Varying the number of input frames affects the success rate and inference speed.}
    \label{fig:ablation_study}
    \vspace{-10pt}
\end{figure}

\noindent\textbf{Ablations on cross-frame decoder designs.} As illustrated in Tab.\ref{table:decoder}, we first study the core design in our DiT-based decoder. For \textit{w/o. cross-atten}, we remove the cross-attention mechanism of our original design and replace it with a direct self-attention network. Our cross-attention design achieves superior overall performance and higher per-task success rates. Notably, self-attention incurs quadratic computational growth with increasing frame numbers, whereas our cross-attention mechanism scales linearly, enabling support for extremely large frame sequences. For \textit{w/o. modulator}, we omit the modulator, treating current (1 frame) and past learnable features (6 frames) indiscriminately without feature splitting and learnable positional embedding, which substantially degrades performance due to the potential dominance of irrelevant historical information over critical current cues. Then, we explore alternative decoder architectures by adopting an \textit{MLP-based decoder} with the same parameter size as our DiT-based model. Due to the large-scale post-training data, the limited representational capability of the MLP leads to a 20\% drop in success rate. The \textit{SiT-based decoder}~\cite{sit} replaces the diffusion objective with a flow-matching one, yielding a lower average score than the DiT-based decoder.

\noindent\textbf{The impact of frame number.}
As shown in Fig.\ref{fig:ablation_study}, we analyze how varying the number of input frames affects different multi-frame modeling strategies. Given that tasks differ in their reliance on temporal information, we conduct a representative analysis on the SimplerEnv benchmark. The study compares CronusVLA 7B, CronusVLA 0.5B, and the Baseline model (post-trained basic single-frame VLA mode) with a naive multi-frame extension. Results show several conclusions: (1) Increasing the total frame number yields an initial increase followed by a subsequent decrease in average success rate. More frames do not consistently lead to better outcomes: CronusVLA 7B performs best with 7 frames, while the CronusVLA 0.5B performs best with 4, suggesting that a moderate amount of temporal information can enhance performance, whereas excessive temporal input may lead to performance degradation. (2) Compared with Baseline, CronusVLA maintains high inference speed across frame counts, avoiding excessive latency overhead. (3) CronusVLA demonstrates strong long-horizon efficiency, with inference speed of baseline models significantly degrading as the horizon extends, while our model remains largely unaffected.

\section{Conclusion}
We presented CronusVLA, a unified framework that efficiently extends single-frame VLA models to the multi-frame paradigm through a two-stage training process. By introducing feature chunking and multi-frame regularization, CronusVLA achieves fast inference, strong performance, and enhanced robustness. Additionally, we proposed SimplerEnv-OR, a benchmark for evaluating observational robustness under diverse disturbances. Extensive experiments across simulated and real-world settings validate the effectiveness and robustness of our approach. Limitations and Future Work are discussed in detail in Appendix.

\bibliography{aaai2026}

\appendix
\let\addcontentsline\oldaddcontentsline

\onecolumn
\begin{center}
  {\LARGE \bf Supplementary Materials }
\end{center}

\begingroup
  \let\addcontentsline\oldaddcontentsline
  \renewcommand{\contentsname}{}

  \begin{center}
  \begin{minipage}{0.9\textwidth}
    \renewcommand{\baselinestretch}{1.5}
    \normalsize
    \setlength{\parskip}{0.6em}
    \setlength{\parindent}{0pt}
    \tableofcontents
  \end{minipage}
  \end{center}
\endgroup

\clearpage
\newpage

\section{SimplerEnv-OR}
\label{sec_details_benchmark}

\subsection{Benchmark Settings}
\begin{figure*}[h]
    \centering
    \includegraphics[width=0.95\linewidth]{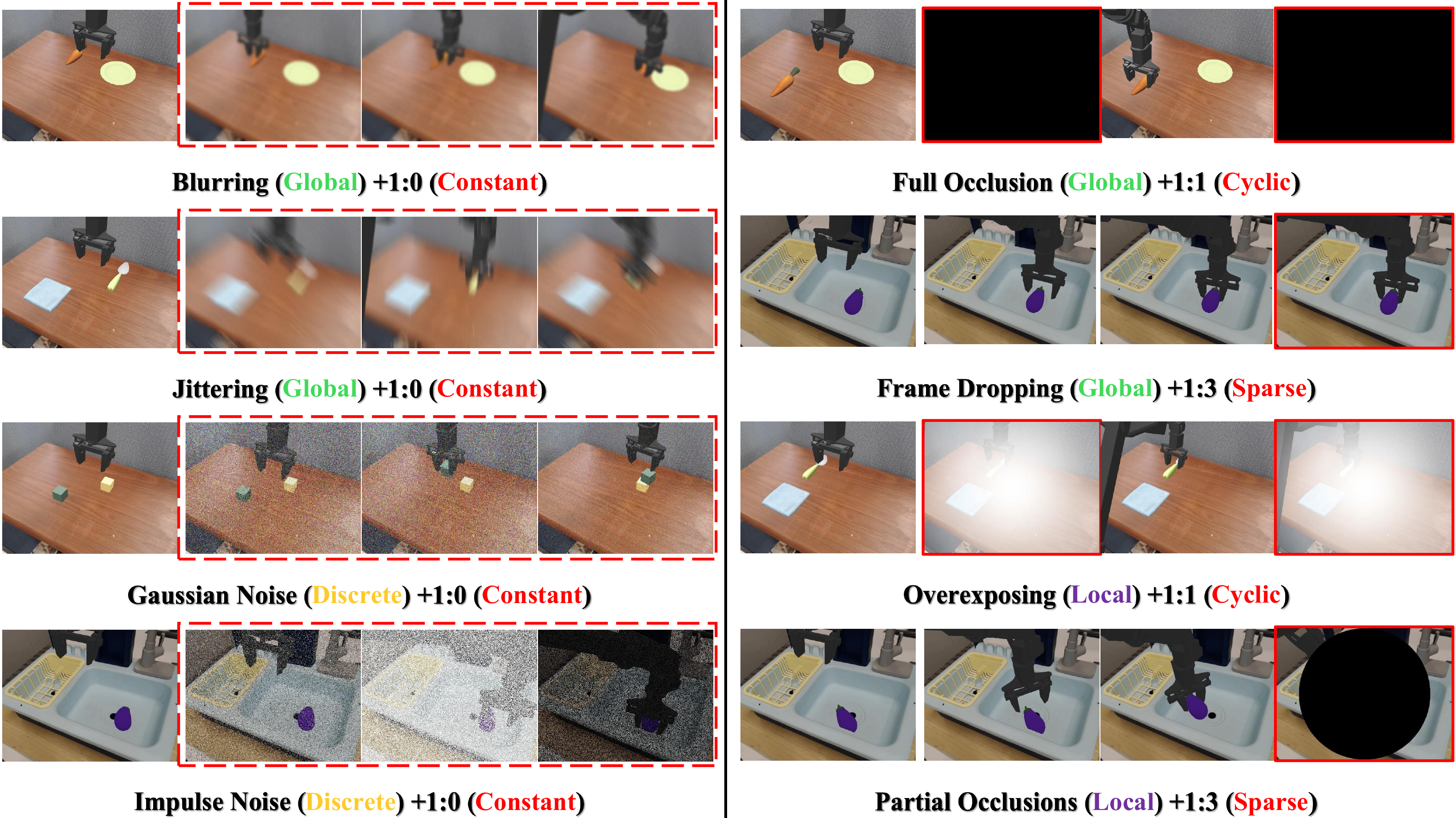}
    \caption{\textbf{SimplerEnv-OR}, which includes different disturbance categories (Global, Local, and Discrete), with different disturbed-to-clean ratios (1:0 for Constant, 1:1 for Cyclic, 1:3 or 1:5 for Sparse), supporting spatial- and temporal-dimension evaluations. }
    \label{fig:interference_summary}
\end{figure*}
SimplerEnv-OR (Observational Robustness) is designed to evaluate model robustness against observational disturbances across temporal and spatial dimensions. It builds upon the WidowX Robot setup from SimplerEnv~\cite{simplerenv}, reusing the original Visual Matching configuration, which includes four subtasks: \textit{Stack Green Cube on Yellow Cube}, \textit{Put Carrot on Plate}, \textit{Put Spoon on Table}, and \textit{Put Eggplant in Basket}. These tasks are all pick-and-place tasks and exhibit stable success rates, making them well-suited for assessing model robustness. As in the original SimplerEnv, these tasks are evaluated in a real-to-sim setting using models trained on the BridgeV2 dataset~\cite{bridgev2}. SimplerEnv-OR directly tests each model under various disturbance types without additional fine-tuning, and computes the average success rate across the four subtasks for each type. We report both the success rate and the predefined robustness score (\textit{R-Socre}) over the corresponding disturbance types. The same random seeds are used to ensure that all baseline models encounter identical disturbances within the same episode and trial during evaluation. A visualization is shown in Fig.\ref{fig:interference_summary}. 

\begin{table*}[h]
\centering
\renewcommand{\arraystretch}{1.2}
\setlength{\tabcolsep}{6pt}
\resizebox{0.75\linewidth}{!}{
\begin{tabular}{l|ccccc}\hline
\toprule
\textbf{Disturbance}& \textbf{Spatial Category}& \textbf{Temporal Category} & \textbf{Ratio Supported}& \textbf{Subtask Number} & \textbf{Trials} \\
\midrule
Blurring &Global&Constant,Cyclic,Sparse& 1:0, 1:1, 1:3& 4&288 \\
Jittering &Global&Constant,Cyclic,Sparse& 1:0, 1:1, 1:3&4&288\\
Frame Dropping &Global&Cyclic,Sparse& 1:1, 1:3, 1:5&4&288 \\
Full Occlusion &Global&Cyclic,Sparse& 1:1, 1:3, 1:5&4&288 \\
Overexposing &Local&Cyclic,Sparse& 1:1, 1:3, 1:5 &4&288 \\
Partial Occlusions &Local&Cyclic,Sparse& 1:1, 1:3, 1:5&4&288 \\
Gaussian Noise &Discrete&Constant,Cyclic,Sparse& 1:0, 1:1, 1:3&4&288 \\
Impulse Noise &Discrete&Constant,Cyclic,Sparse& 1:0, 1:1, 1:3 &4&288 \\\rowcolor{gray!15}
None (Origin) &-&-& 0:1 &4&96\\\bottomrule
\hline
\end{tabular}}
\caption{\textbf{Task details of SimplerEnv-OR,} including types of observational disturbances, corresponding spatial categories, temporal categories, supported ratios (i.e., Disturbed:Clean, frame number), subtask number, and evaluation trials number.}
\label{tab:SimplerEnv-OR_summary}
\end{table*}

\begin{table*}[h]
\centering
\renewcommand{\arraystretch}{1.2}
\setlength{\tabcolsep}{6pt}
\resizebox{0.95\linewidth}{!}{
\begin{tabular}{l|ccc}
\hline\toprule
\textbf{Disturbance}& \textbf{Foundation} & \textbf{Core Parameters} & \textbf{Sampling Configurations} \\
\midrule
Blurring & gaussian blur & kernel size & size$\in\{(11,11), (15,15), (29,29)\}$, $\sigma \in \{(5,0), (0,5)\}$ \\
Jittering & motion blur & direction, kernel size & direction  $\in\{H, V, D\}$, size $\in\{15, 25, 50\}$\\
Frame Dropping & frame replication & temporal offset & use \texttt{past\_image\_list}[-1] \\
Full Occlusion & full mask & (none) & image $\gets 0$ \\
Overexposing &exposure varying&position, intensity, radius & intensity  $\in\{0.7, 0.85, 1.0\}$, radius based on image size \\
Partial Occlusions &partial mask & position, radius & $ \{1/4, 2/4, 3/4\}$width, radius based on image size \\
Gaussian Noise &gaussian noise & coordinates, std & std $\in\{25, 50, 75\}$ \\
Impulse Noise & salt and pepper noise&coordinates, amount, salt\_vs\_pepper & amount $\in\{0.2, 0.5, 0.8\}$, s/p $\in\{0.0, 0.5, 1.0\}$ \\\bottomrule\hline
\end{tabular}}
\caption{\textbf{Detailed configurations of disturbances,} summary of visual disturbance types, foundation tools, core parameters, and sampling configurations.}
\label{tab:disturbance_summary}
\end{table*}

A detailed summary of disturbances conducted in our SimplerEnv-OR is illustrated in Tab.\ref{tab:SimplerEnv-OR_summary} and Tab.\ref{tab:disturbance_summary}. Notably, commonly used image augmentations during VLA models training, such as random resized cropping, adjustments to brightness, contrast, saturation, and hue, are beyond the scope of our benchmark, and the designed tasks in SimperEnv-OR cannot be directly solved through these simple data augmentation techniques.

\subsection{Spatial Evaluations on SimplerEnv-OR}

\begin{figure*}[h]
    \centering
    \includegraphics[width=0.95\linewidth]{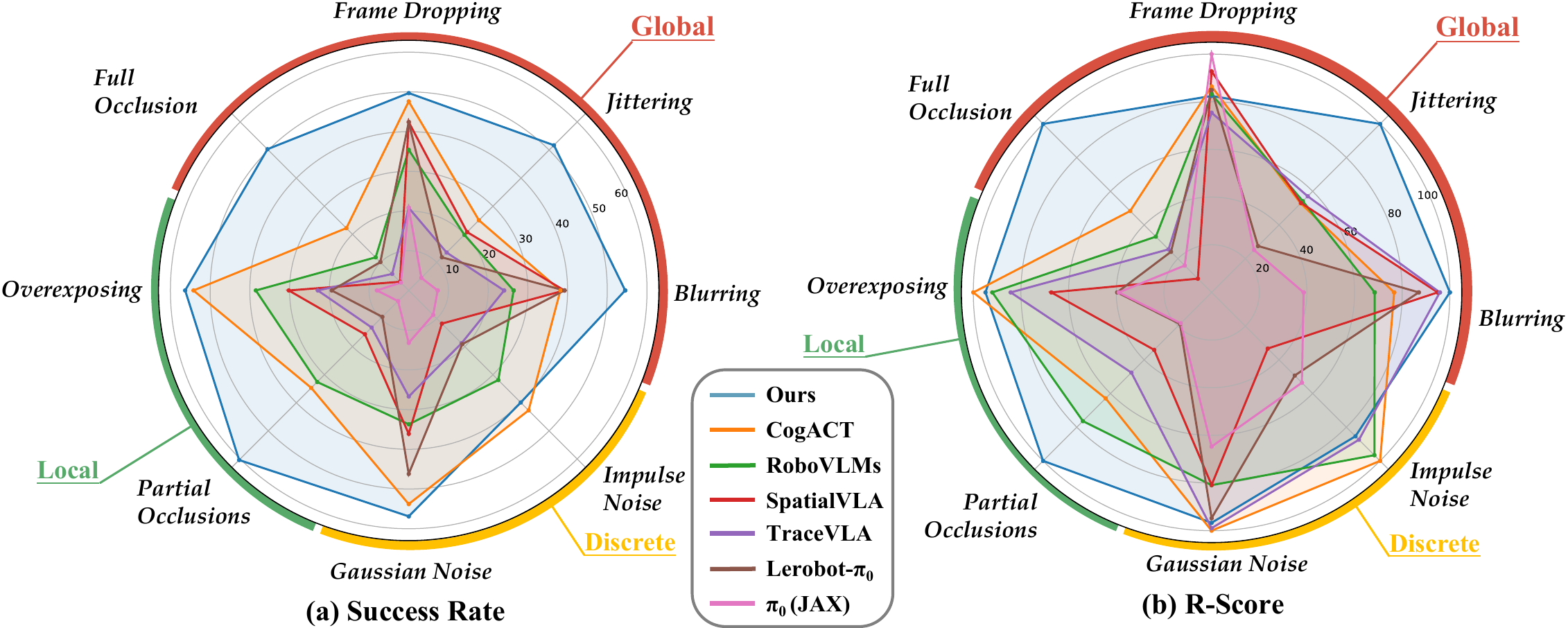}
    \caption{\textbf{Spatial-dimension results on SimplerEnv-OR.} The left panel presents the average success rates across all trials (under 8 major disturbance types, 3 spatial categories). The right panel shows the corresponding R-Score performances.}
    \label{fig:spatial_evaluation}
\end{figure*}

The spatial dimension assesses model robustness by injecting various types of disturbances into clean observations, simulating realistic interference encountered by cameras in practical deployments. Observation disturbances are typically categorized into three categories: \textbf{Global}, \textbf{Local}, and \textbf{Discrete}. \textbf{Global} disturbances affect the entire observation, resulting in complete information loss (e.g., Full Occlusion) or uniform degradation (e.g., Blurring). \textbf{Local} disturbances are confined to specific regions of the observation, leaving the remaining areas unaffected, such as Partial Occlusions. \textbf{Discrete} disturbances refer to randomly distributed, non-contiguous noise patterns where only a subset of pixels within a region are affected, characterizing sparse and irregular corruption.

As illustrated in Fig.~\ref{fig:spatial_evaluation}, the left and right panels show the average success rates and R-Scores, respectively, across Global, Local, and Discrete in SimplerEnv-OR. Our CronusVLA achieves the highest success rate under nearly all disturbance types. It also demonstrates superior robustness in R-Score, particularly under Full Occlusion, Partial Occlusion, Jittering, and Blurring, while maintaining competitive performance across other types. This robustness can be attributed to the temporal modeling of our multi-frame method. SpatialVLA shows strong robustness to Frame Dropping and Blurring but is highly sensitive to other disturbances, especially Full Occlusion, likely due to its reliance on zero-shot depth estimation. Although RoboVLMs and TraceVLA perform less effectively than SpatialVLA in the original SimplerEnv, they exhibit enhanced robustness on SimplerEnv-OR. CogACT achieves the best R-Score performance on several disturbance types but falls significantly behind CronusVLA in Full Occlusion, Partial Occlusion, Blurring and Jittering. We also test both the JAX and the Lerobot version of $\pi_0$.

\begin{figure}[htbp]
    \centering
    \includegraphics[width=0.9\linewidth]{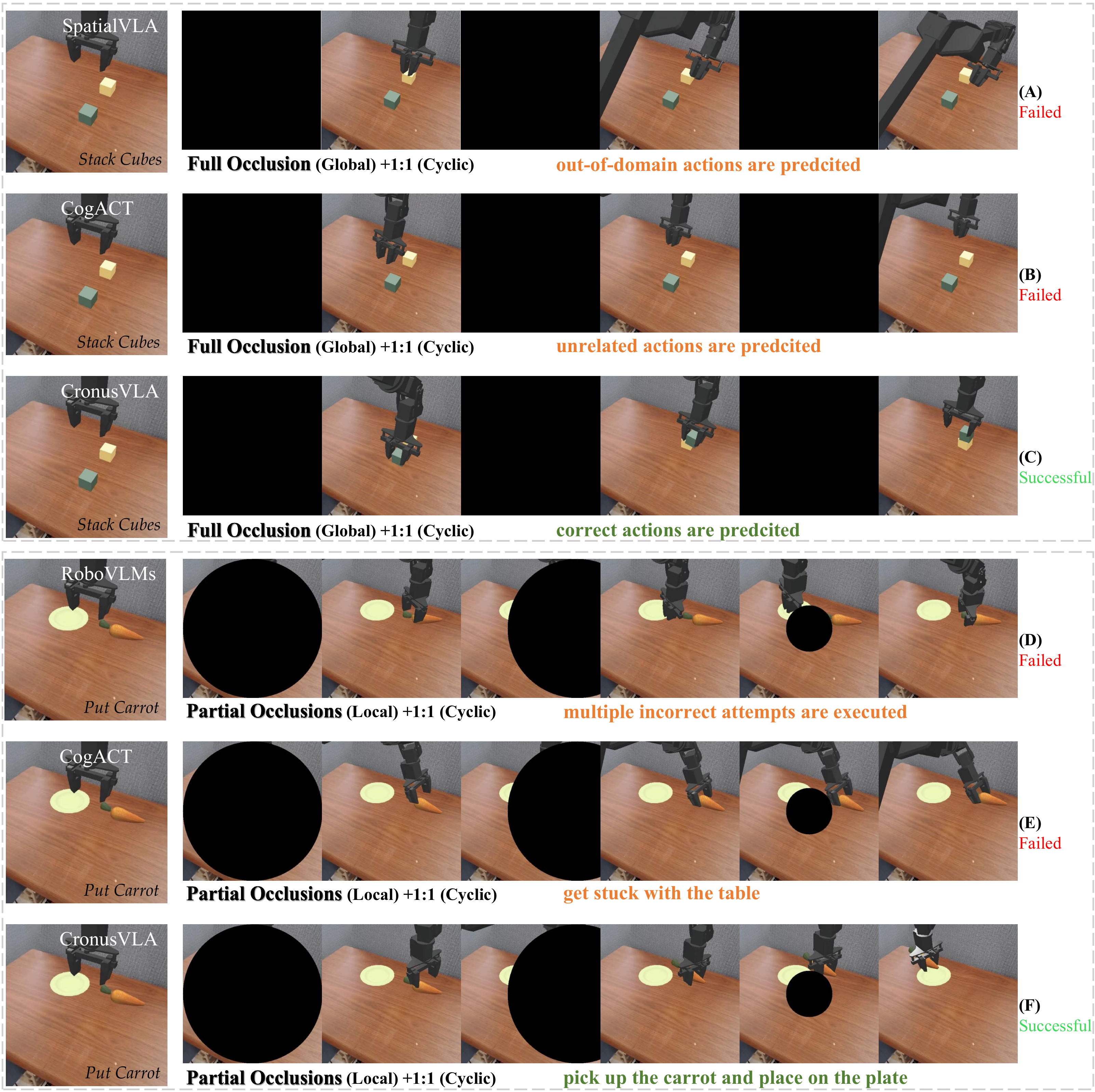}
    \caption{\textbf{Visualizations of spatial-dimension testing on SimplerEnv-OR.} Qualitative comparisons under spatial disturbances such as Cyclic Global Full Occlusion (upper) and Cyclic Local Partial Occlusion (bottom).}
    \label{fig:spatial_visual}
    \vspace{1em}

    \renewcommand{\arraystretch}{1.2}
    \setlength{\tabcolsep}{3.5pt}
    \resizebox{0.9\linewidth}{!}{
    \begin{tabular}{l|ccc|cccc|cccc|ccc}
    \toprule
    \multirow{2}{*}{Method} 
    & \multicolumn{3}{c|}{Constant (1:0)} 
    & \multicolumn{4}{c|}{Cyclic (1:1)} 
    & \multicolumn{4}{c|}{Sparse (1:3)} 
    & \multicolumn{3}{c}{Sparse (1:5)} \\
    \cmidrule(lr){2-4} \cmidrule(lr){5-8} \cmidrule(lr){9-12} \cmidrule(lr){13-15}
    & Global-p & Discrete & Avg. 
    & Global & Discrete & Local & Avg. 
    & Global & Discrete & Local & Avg. 
    & Global-p & Local & Avg. \\
    \midrule
    $\pi_0$(jax) & 7.3& 10.9& 9.1& 8.3& 11.5& 2.6& 7.7& 7.6& 10.4& 3.7& 7.3& 14.1& 11.5& 12.8\\
    lerobot-$\pi_0$ & 27.1& 32.8& 30.0& 21.6& 31.8& 6.8& 20.4& 22.9& 33.3& 8.3& 21.9& 39.1& 28.1& 33.6\\
    TraceVLA & 11.5& 21.4& 16.4& 16.4& 22.4& 14.1& 17.3& 20.1& 24.5& 21.9& 21.6& 12.0& 18.2& 15.1\\
    SpatialVLA & 22.9 & 18.2 & 20.6 & 25.8 & 21.9 & 14.6 & 22.0 & 28.9 & 31.8 & 26.0 & 28.9 & 26.0 & 28.1 & 27.1 \\
    RoboVLMs   & 16.7 & 24.0 & 20.3 & 20.3 & 31.8 & 25.5 & 24.5 & 26.6 & 42.7 & 38.5 & 33.6 & 29.7 & 42.7 & 36.2 \\
    CogACT     & 17.2 & \textbf{41.7} & 29.4 & 28.4 & 52.1 & 37.0 & 36.5 & 41.2 & 51.0 & 43.8 & 44.3 & 43.2 & 52.6 & 47.9 \\
    \rowcolor{gray!15}
    Ours  & \textbf{38.5} & 35.4 & \textbf{37.0} & \textbf{49.5} & \textbf{53.1} & \textbf{57.3} & \textbf{52.3} & \textbf{58.1} & \textbf{56.8} & \textbf{59.4} & \textbf{58.1} & \textbf{55.7} & \textbf{58.3} & \textbf{57.0} \\
    \bottomrule
    \end{tabular}
    }
    \captionof{table}{\textbf{Spatial-dimension success rates under different ratios.} For each disturbed-to-clean ratio (from 1:0 to 1:5), we report all success rates. \textit{Global-p} indicates that a part of the Global disturbances are included.}
    \label{tab:detailed_perturbation_sr}
\end{figure}

\begin{table}[h]
\centering
\renewcommand{\arraystretch}{1.2}
\setlength{\tabcolsep}{3.5pt}
\resizebox{0.9\linewidth}{!}{
\begin{tabular}{l|ccc|cccc|cccc|ccc}
\toprule
\multirow{2}{*}{Method} 
& \multicolumn{3}{c|}{Constant (1:0)} 
& \multicolumn{4}{c|}{Cyclic (1:1)} 
& \multicolumn{4}{c|}{Sparse (1:3)} 
& \multicolumn{3}{c}{Sparse (1:5)} \\
\cmidrule(lr){2-4} \cmidrule(lr){5-8} \cmidrule(lr){9-12} \cmidrule(lr){13-15}
& Global-p & Discrete & Avg. 
& Global & Discrete & Local & Avg. 
& Global & Discrete & Local & Avg. 
& Global-p & Local & Avg. \\
\midrule
$\pi_0$(jax) &34.9& 52.3& 43.6& 39.9& 54.8& 12.5& 36.8& 36.1& 49.8& 17.4& 34.9& 67.3& 54.8& 61.1\\
lerobot-$\pi_0$ & 54.2& 65.6& 59.9& 43.2& 63.5& 13.5& 40.9& 45.8& 66.7& 16.7& 43.8& 78.1& 56.2& 67.2\\
TraceVLA & 41.4& \textbf{77.1}& 59.2& 59.2& 80.9& 50.8& 62.5& 72.4& 88.4& 79.0& 78.0& 43.3& 65.8& 54.5\\
SpatialVLA & 50.0 & 39.8 & 44.9 & 56.3 & 47.8 & 31.8 & 48.1 & 63.1 & 69.4 & 56.9 & 63.1 & 56.9 & 61.4 & 59.1 \\
RoboVLMs   & 39.0 & 56.1 & 47.6 & 47.6 & 74.4 & 59.8 & 57.3 & 62.2 & \textbf{100.0} & 90.3 & 78.7 & 69.5 & \textbf{100.0} & 84.8 \\
CogACT     & 31.1 & 75.5 & 53.3 & 51.4 & \textbf{94.4} & 67.0 & 66.1 & 74.5 & 92.5 & 79.3 & 80.2 & 78.3 & 95.3 & 86.8 \\
\rowcolor{gray!15}
Ours  & \textbf{63.8} & 58.6 & \textbf{61.2} & \textbf{81.9} & 88.0 & \textbf{94.9} & \textbf{86.7} & \textbf{96.2} & 94.0 & \textbf{98.3} & \textbf{96.2} & \textbf{92.3} & 96.6 & \textbf{94.4} \\
\bottomrule
\end{tabular}
}
\caption{\textbf{Spatial-dimension R-Score under different ratios.} For each disturbed-to-clean ratio (from 1:0 to 1:5), we report the R-Score corresponding to different spatial-level categories. \textit{Global-p} indicates that a part of the Global disturbances are included.}
\label{tab:detailed_perturbation_rs}
\end{table}

Fig.~\ref{fig:spatial_visual} presents qualitative results. In the upper, \textit{Stack Green Cube on Yellow Cube} task under \textbf{Global} (Full Occlusion) and Cyclic disturbances, SpatialVLA frequently outputs out-of-domain actions and feature representation, leading to task failures. CogACT, also relying solely on the current frame, produces contextually irrelevant actions when occlusion occurs. In contrast, CronusVLA, equipped with multi-frame modeling, demonstrates superior robustness under such conditions. In the bottom, \textit{Put Carrot on Plate} task under \textbf{Local} (Partial Occlusion) and Cyclic disturbances, RoboVLMs, although leveraging multi-frame modeling, depend heavily on accurate temporal information. When this information is compromised, the model avoids out-of-distribution actions but repeats ineffective attempts. CogACT misinterprets the scene, interacting with irrelevant objects (e.g., the table), resulting in task failure. CronusVLA maintains robust performance under Partial Occlusion by effectively integrating temporal context. 

We report more detailed experimental results in Tab.\ref{tab:detailed_perturbation_sr} and Tab.\ref{tab:detailed_perturbation_rs}. All values represent the average performance across all disturbance types within the respective spatial category. Our model demonstrates clear advantages over others in both Global and Local evaluations, and exhibits strong robustness in the Discrete setting.

\subsection{Temporal Evaluations on SimplerEnv-OR}
The temporal dimension assesses model robustness by injecting disturbances at varying intervals, simulating different disturbance ratios in otherwise clean observations. This design captures realistic disturbance patterns encountered by cameras in practical deployment scenarios, ranging from persistent to intermittent interference caused by external factors. To enable a meaningful qualitative analysis, we evaluate three representative settings: Constant (1:0), Cyclic (1:1), and Sparse (1:3, 1:5).

Firstly, to better illustrate the impact of different disturbance ratios on model performance, we present detailed scores under Full Occlusion in Tab.\ref{tab:full_occlusion_summary}, along with results from the original SimplerEnv across all four subtasks. As shown in this table, success rates generally improve as the disturbance ratio decreases. Under Cyclic disturbances, most models exhibit near-zero success rates, as alternating normal and fully occluded frames severely disrupt single-frame decision logic, often resulting in out-of-distribution actions. In contrast, our CronusVLA, equipped with temporal modeling, effectively leverages past observations to infer correct decisions, demonstrating superior robustness.

\begin{table}[h]
\centering
\renewcommand{\arraystretch}{1.2}
\setlength{\tabcolsep}{4pt}
\vspace{2mm}
\resizebox{0.9\linewidth}{!}{
\begin{tabular}{l|cccccc|cccccc}
\toprule
\multirow{2}{*}{Method} 
& \multicolumn{6}{c|}{Cyclic (1:1)} 
& \multicolumn{6}{c}{Sparse (1:3)} \\
\cmidrule(lr){2-7} \cmidrule(lr){8-13}
& Spoon & Carrot & Blocks & Eggplant & Avg. & R-Score 
& Spoon & Carrot & Blocks & Eggplant & Avg. & R-Score \\
\midrule
SpatialVLA  & 0.0  & 0.0  & 0.0  & 0.0  & 0.0  & 0.0  & 0.0  & 0.0  & 0.0  & 0.0  & 0.0  & 0.0  \\
RoboVLMs    & 0.0  & 0.0  & 0.0  & 0.0  & 0.0  & 0.0  & 41.7 & 12.5 & 0.0  & 4.2  & 14.6 & 30.5 \\
CogACT      & 8.3  & 0.0  & 0.0  & 8.3  & 4.2  & 5.4  & 25.0 & 25.0 & 4.2  & 54.2 & 27.1 & 44.2 \\
\rowcolor{gray!15}
Ours   & \textbf{45.8} & \textbf{25.0} & \textbf{4.2}  & \textbf{66.7} & \textbf{35.4} & \textbf{50.4} & \textbf{58.3} & \textbf{66.7} & \textbf{8.3}  & \textbf{95.8} & \textbf{57.3} & \textbf{86.6} \\
\midrule
\multirow{2}{*}{}
& \multicolumn{6}{c|}{Sparse (1:5)} 
& \multicolumn{6}{c}{\textit{WR-VM (SimplerEnv)}} \\
\cmidrule(lr){2-7} \cmidrule(lr){8-13}
& Spoon & Carrot & Blocks & Eggplant & Avg. & R-Score 
& \textit{Spoon} & \textit{Carrot} & \textit{Blocks} & \textit{Eggplant} & \textit{Avg.} & \textit{-} \\
\midrule
SpatialVLA  & 0.0  & 8.3  & 0.0  & 29.2  & 9.4  & 14.3 & 20.8 & 37.5 & 41.7 & 83.3  & 45.8 & -- \\
RoboVLMs    & 33.3 & 20.8 & 0.0  & 29.2  & 20.8 & 39.3 & 50.0 & 37.5 & 0.0  & 83.3  & 42.7 & -- \\
CogACT      & 37.5 & 29.2 & 0.0  & 75.0  & 35.4 & 50.8 & \textbf{75.0} & 50.0 & 16.7 & 79.2  & 55.2 & -- \\
\rowcolor{gray!15}
Ours   & \textbf{75.0} & \textbf{37.5} & \textbf{20.8} & \textbf{100.0} & \textbf{58.3} & \textbf{95.4} & 66.7 & \textbf{54.2} & \textbf{20.8} & \textbf{100.0} & \textbf{60.4} & -- \\
\bottomrule
\end{tabular}
}
\caption{\textbf{Detailed experimental results under Full Occlusion} across three disturbance ratios: 1:1, 1:3, 1:5. Origin results of SimplerEnv WR-VM setting is also reported. Success rates (\%) are reported. “R-Score” indicates the robustness score.}
\label{tab:full_occlusion_summary}
\end{table}

\begin{figure*}[h]
    \centering
    \includegraphics[width=0.9\linewidth]{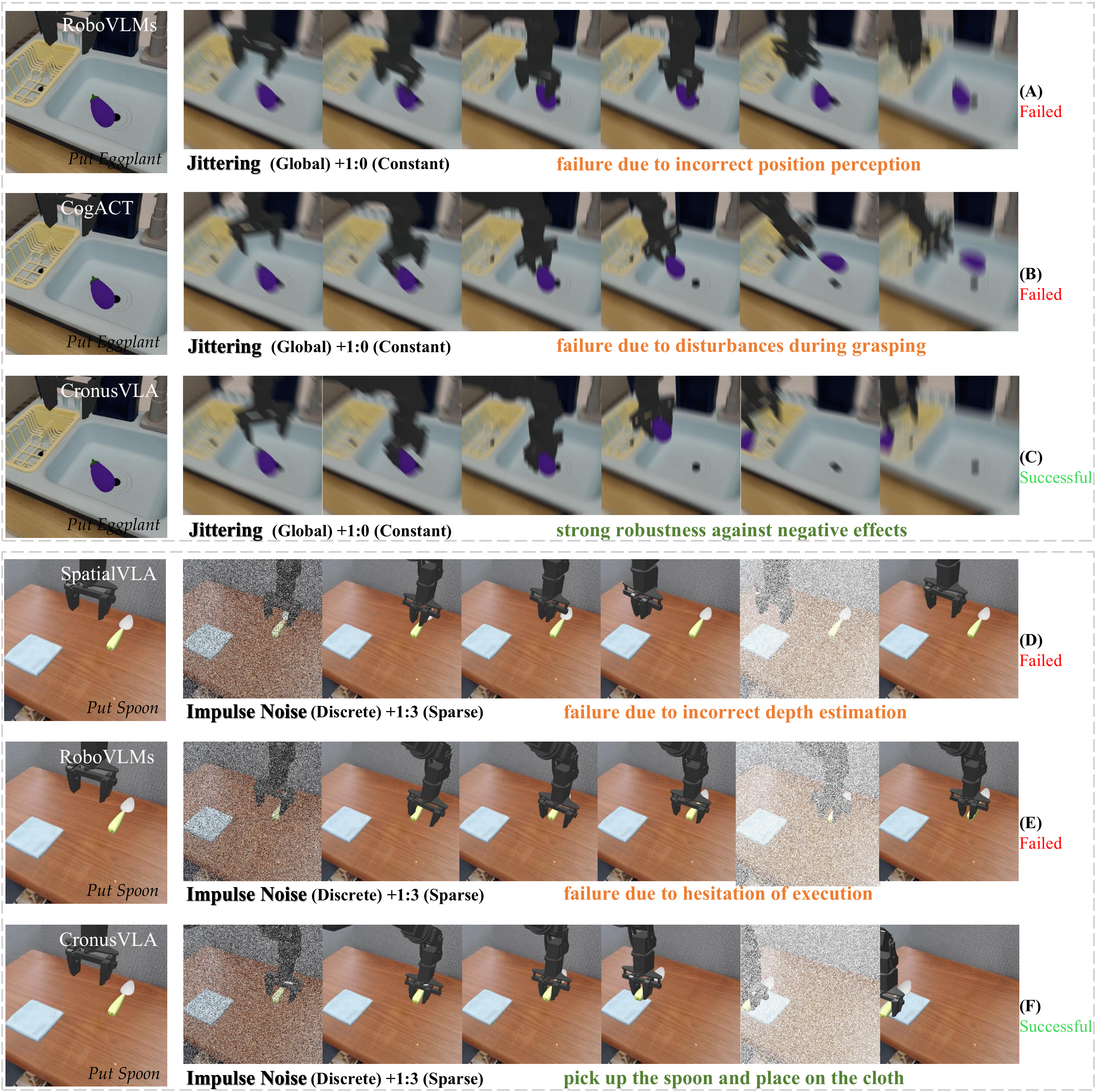}
    \caption{\textbf{Visualizations of temporal-dimension testing on SimplerEnv-OR.} Qualitative comparisons under spatial disturbances such as Constant Global Jittering (upper) and Sparse Discrete Impulse Noise (bottom).}
    \label{fig:temporal_visual}
\end{figure*}

We present qualitative results in Fig.\ref{fig:temporal_visual}, including a Constant disturbance under Jittering and a Sparse disturbance under Impulse Noise. In the upper example, Constant disturbances significantly impact model performance: RoboVLMs exhibit incorrect position perception due to insufficient accurate historical information, which may be caused by its LSTM-based design. CogACT fails to maintain object grasping under disturbance. In the bottom example, SpatialVLA suffers from wrong depth estimations, resulting in failed grasps at incorrect heights, while RoboVLMs show execution hesitation, reducing success rates. In contrast, CronusVLA demonstrates higher success rates under extreme conditions (Constant) and remains largely unaffected under mild disturbances (Sparse).

\begin{figure*}[h]
    \centering
    \includegraphics[width=0.95\linewidth]{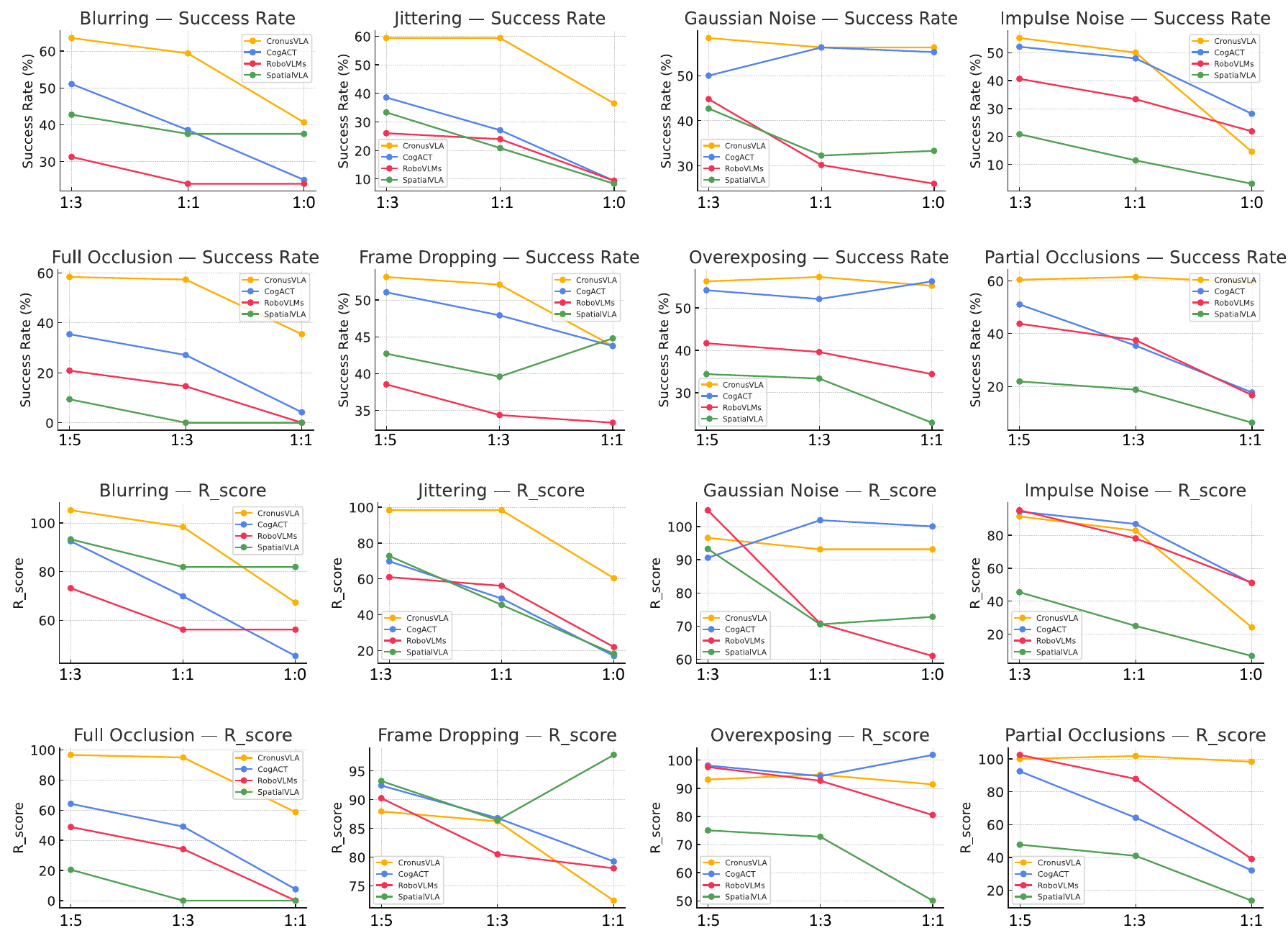}
    \caption{\textbf{Impact of disturbance types and ratios on models performance}. Both the success rate and R-Score are reported.}
    \label{fig:interference_summary_3}
\end{figure*}

Fig.~\ref{fig:interference_summary_3} illustrates the impact of varying types and ratios of disturbances on different models. Some disturbance types exhibit increased success rates at higher disturbance ratios. We attribute this to two main factors: (1) inherent randomness in both the models and disturbances, as well as task-specific variance, which makes performance more stable when averaged over more trials; (2) disturbances during critical moments may disproportionately affect model performance, and higher ratios do not necessarily increase interference at those moments. SimplerEnv-OR currently adopts fixed-frequency disturbances, which could be improved by incorporating targeted perturbations on keyframes for more accurate robustness evaluation.

\FloatBarrier
\subsection{More Discussions}
Our design features three key properties:

\textbf{(1) Plug-and-Play and Transferable.} We focus on evaluating the robustness of observations, which are typically obtained via environment rendering. To ensure decoupling from simulation environments, we introduce disturbances directly on rendered images (as NumPy arrays), without modifying the simulator itself. This approach is compatible with any simulation or standard benchmark.

\textbf{(2) Extensibility.} The types and intensities of disturbances are fully customizable without additional configuration, allowing seamless integration of various image processing techniques.

\textbf{(3) Reproducibility.} We ensure consistency across testing trials by deterministically mapping each trials id to a random seed using its hash value, guaranteeing identical disturbances for different models during evaluation.

However, our approach still has several limitations, such as the incomplete coverage of disturbance types, which we leave for future work. Expanding the number of baselines, supporting more simulation environments, and incorporating more accurate and comprehensive evaluation protocols are also our future directions.

We hope to inspire the community to consider robustness evaluation in Vision-Language-Action (VLA) models, which is crucial for VLA real-world deployment and future applications.

\begin{table}[h]
\centering
\renewcommand{\arraystretch}{1.2}
\resizebox{0.7\linewidth}{!}{
\begin{tabular}{lcccccccc}
\toprule
& \multicolumn{6}{c}{Temporal Dimension} & \multicolumn{2}{c}{{\textit{WidowX Robot}}} \\
& \multicolumn{2}{c}{Constant (1:0)} & \multicolumn{2}{c}{Cyclic (1:1)} & \multicolumn{2}{c}{Sparse (1:3)} & \multicolumn{2}{c}{{\textit{VM}}} \\
\cmidrule(lr){2-7}
Methods & R-Score & SR & R-Score & SR & R-Score & SR & - & \textit{SR} \\
\midrule
$\pi_0$(jax)~\cite{pi0}  & 43.5 & 9.1 & 36.8 & 7.7 & 34.9 & 7.3  & - & \textit{20.9} \\
lerobot-$\pi_0$~\cite{cadene2024lerobot}  & 59.9 & 30 & 40.9 & 20.4 & 43.8 & 21.9  & - & \textit{50.0} \\
TraceVLA~\cite{Tracevla}  & 59.2 & 16.4 & 62.5 & 17.3 & 78.0 & 21.6  & - & \textit{27.7} \\
RoboVLMs~\cite{robovlms}     & 47.6 & 20.3 & 57.3 & 24.5 & 78.7 & 33.6  & - & \textit{42.7} \\
SpatialVLA~\cite{spatialvla}   & 44.9 & 20.6 & 48.0 & 22.0 & 63.1 & 28.9 & - & \textit{45.8} \\
CogACT~\cite{cogact}       & 53.3 & 29.4 & 66.1 & 36.5 & 80.2 & 44.3 & - & \textit{55.2} \\\midrule
\rowcolor{gray!15}
\textbf{Ours (7B)} & \textbf{61.2} & \textbf{37.0} & \textbf{86.7} & \textbf{52.3} & \textbf{96.2} & \textbf{58.1} & - & \textit{60.4} \\
\midrule
\midrule
& \multicolumn{6}{c}{Spatial Dimension} & \multicolumn{2}{c}{\multirow{2}{*}{Total Avg.}} \\
& \multicolumn{2}{c}{Global} & \multicolumn{2}{c}{Local} & \multicolumn{2}{c}{Discrete} & & \\
\cmidrule(lr){2-7}
Methods & R-Score & SR & R-Score & SR & R-Score & SR & \textbf{R-Score} & SR \\
\midrule
$\pi_0$(jax)~\cite{pi0} & 42.6 & 8.9 & 28.2 & 5.9 & 52.1 & 10.9  & 41.1 & 8.6 \\
lerobot-$\pi_0$~\cite{cadene2024lerobot} & 51.7 & 25.9 & 28.8 & 14.4 & 65.3 & 32.6  & 49.4 & 24.7 \\
TraceVLA~\cite{Tracevla}  & 58.1 & 16.1 & 65.3 & 18.1 & 81.9 & 22.7  & 65.8 & 18.2 \\
RoboVLMs~\cite{robovlms}     & 54.7 & 23.4 & 83.3 & 35.6 & 76.8 & 32.8 & 67.4 & 28.8 \\
SpatialVLA~\cite{spatialvla}   & 57.6 & 26.4 & 50.0 & 22.9 & 52.4 & 24.0 & 54.4 & 24.9 \\
CogACT~\cite{cogact}      & 60.2 & 33.2 & 80.5 & 44.4 & \textbf{87.4} & 48.3 & 72.1 & 39.8 \\\midrule
\rowcolor{gray!15}
\textbf{Ours (7B)} & \textbf{85.4} & \textbf{51.6} & \textbf{96.6} & \textbf{58.3} & 80.2 & \textbf{48.4} & \textbf{86.9} & \textbf{52.4} \\
\bottomrule
\end{tabular}
}

\caption{\textbf{Additional results on SimplerEnv-OR.} Top: Temporal dimension with different frequencies. Bottom: Spatial dimension with different patterns. R-Score indicates the predefined robustness score. SR denotes success rate. $\pi_0$ models on different frameworks (JAX or Torch) are included.}

\label{tab:detailed_wrvm_cleaned}
  \vspace{-10pt}
\end{table}

\clearpage
\newpage

\section{Case Study for Long-horizon Tasks}
\label{sec_Case_Study}

\subsection{Simulation Experiments} 
\begin{figure*}[h]
    \centering
    \includegraphics[width=1\linewidth]{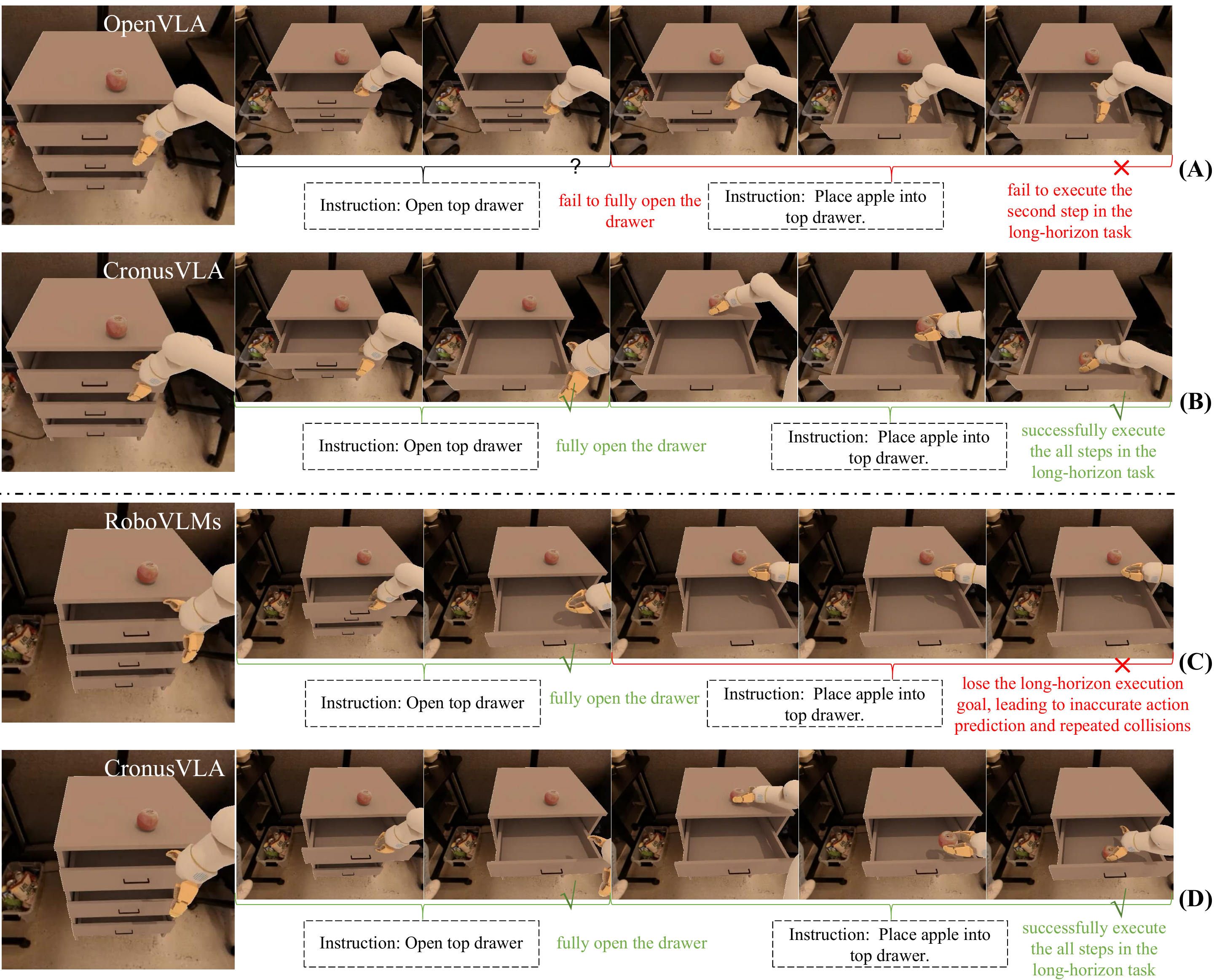}
    \caption{\textbf{Case study for \textit{Put in Drawer} in SimplerEnv Benchmark.} We compare and analyze the same trial performed by CronusVLA and OpenVLA in (A) and (B), as well as a trial performed by CronusVLA and RoboVLMs in (C) and (D).}
    \label{fig:case_study_put_in_drawer}
\end{figure*}

Our model achieves a considerable performance on the \textit{Put in Drawer} task of SimplerEnv~\cite{simplerenv} benchmark, reaching 64.8\% and 65.1\% in Visual Matching and Variant Aggregation, compared to baseline methods including OpenVLA~\cite{openvla}, SpatialVLA~\cite{spatialvla}, RoboVLMs~\cite{robovlms} and TraceVLA~\cite{Tracevla}, which achieve only 0\%–30\% success rates. As shown in Fig.\ref{fig:case_study_put_in_drawer}, we present a case study in the SimplerEnv setting with a Google Robot. In (A) and (B), OpenVLA struggles to fully open the drawer, blocking progress at the first step. In contrast, CronusVLA consistently completes all steps of the long-horizon task. In (C) and (D), although RoboVLMs can successfully open the top drawer initially, they fail to maintain the new execution goal, resulting in inaccurate actions and repeated collisions. This may stem from inadequate modeling capacity to adapt execution goals throughout long-horizon tasks. CronusVLA, by contrast, executes all steps smoothly and reliably.

\subsection{Real-world Experiments} 
\begin{figure*}[h]
    \centering
    \includegraphics[width=1\linewidth]{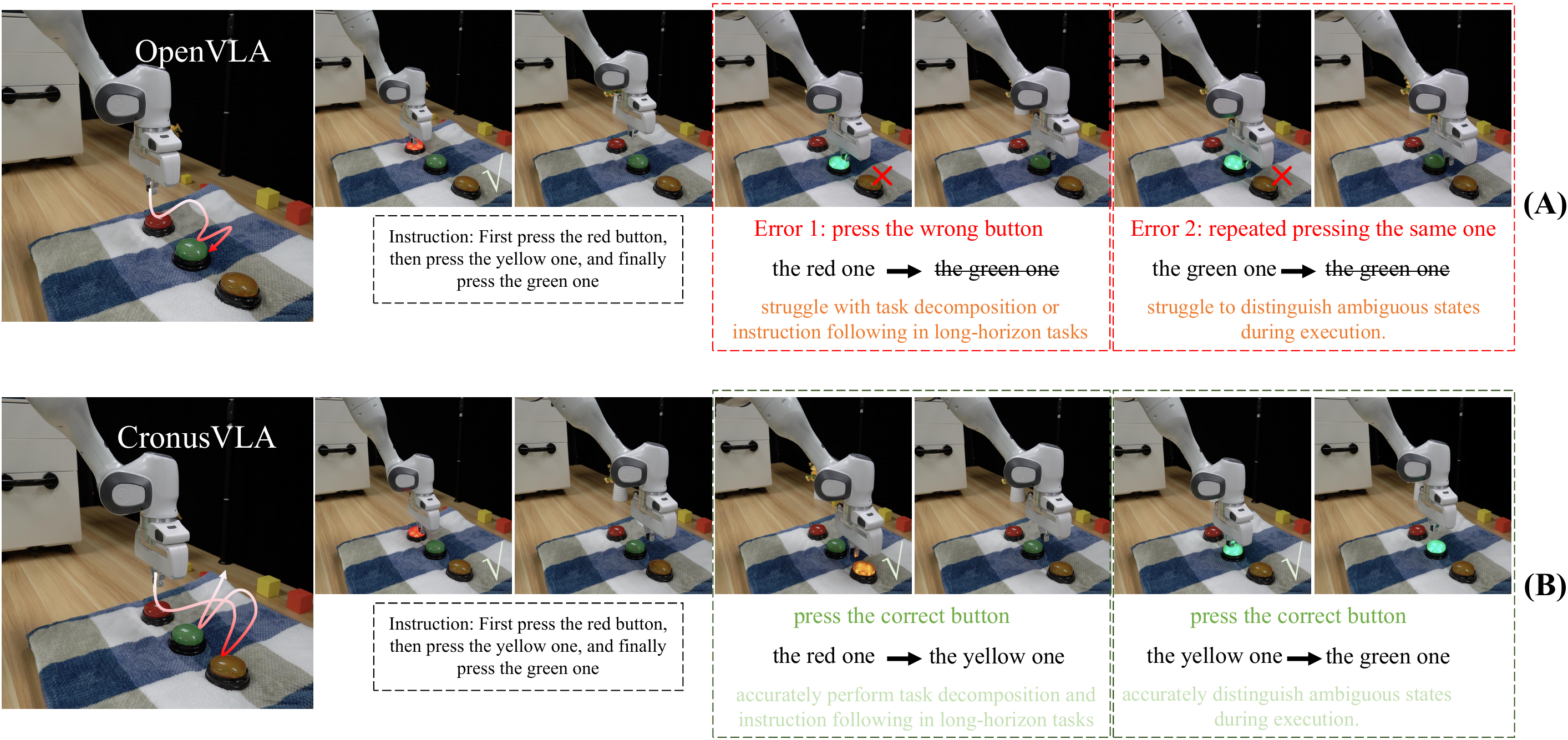}
    \caption{\textbf{Case study for \textit{Press the buttons in order} in our real-world long-horizon tasks.} We compare and analyze the same trial performed by CronusVLA and OpenVLA in (A) and (B).}
    \label{fig:case_study_press_in_order}
\end{figure*}

Our CronusVLA adopts a multi-frame modeling strategy to improve the performance of VLA models, particularly in long-horizon tasks. We find that incorporating temporal information from multiple frames considerably enhances the model’s ability to perform multi-step execution, which is crucial for accurately completing complex, sequential instructions. As illustrated in Fig.\ref{fig:case_study_press_in_order}, we present a representative case from a real-world long-horizon task: pressing red, yellow, and green buttons in sequence. This task requires both instruction following and proper task decomposition. OpenVLA, as a single-frame model, struggles with these aspects and fails to execute the correct order, as shown in Fig.\ref{fig:case_study_press_in_order} (A, Error 1). Furthermore, due to the presence of ambiguous states, where similar observations occur before and after button presses, OpenVLA cannot reliably distinguish between them, leading to repeated presses of the same button (A, Error 2). In contrast, CronusVLA demonstrates strong task decomposition and instruction-following capabilities, and its multi-frame temporal modeling provides critical context for resolving such ambiguities, enabling more robust and accurate execution in long-horizon scenarios.

\clearpage
\newpage

\section{Additional Experiments}
\label{sec_Additional_Experiments}
\subsection{More Explorations on Multi-frame Training}
\label{sec_Additional_Experiments_More_Explorations_on_Multi_frame_Training}
\begin{figure}[h]
    \centering
    \includegraphics[width=0.5\linewidth]{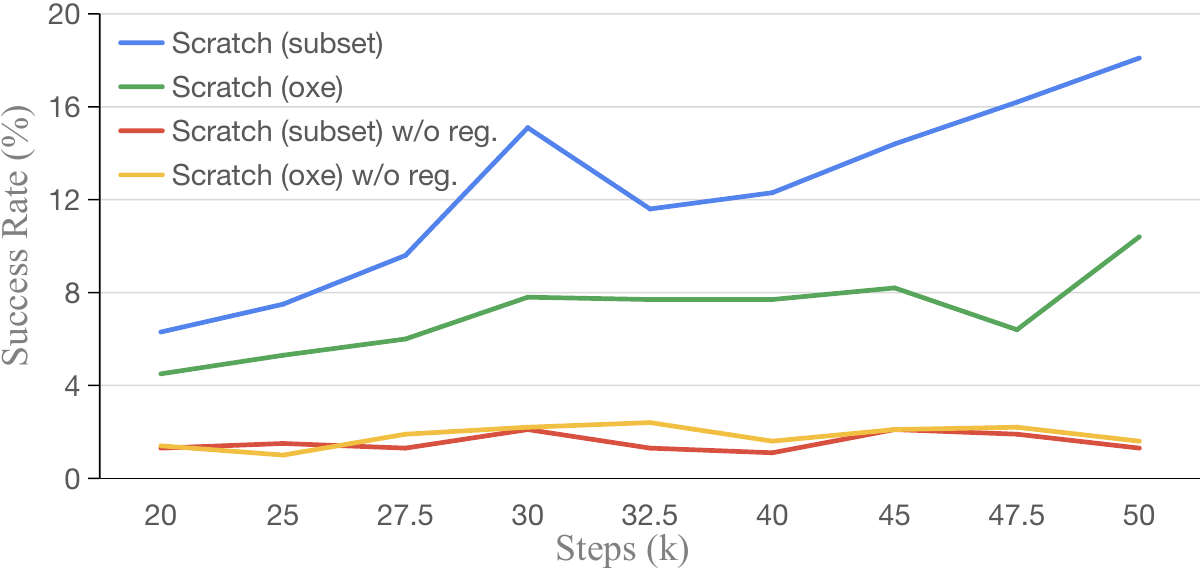}
    \caption{\textbf{Single-frame pretraining and multi-frame regularization accelerate the convergence of multi-frame training.} Convergence trends of training multi-frame VLA are shown, based on CronusVLA 7B (6 past frames) and SimplerEnv.}
    \label{fig:pretraining_importance}
\end{figure}

\noindent\textbf{Single-frame pretraining considerably accelerates the convergence of multi-frame training.} We compare the effectiveness of applying multi-frame post-training to a pretrained single-frame VLA model with training a multi-frame VLA entirely from scratch. To explore the differences under equal GPU budgets, we adopt identical settings and train a plain VLM model (without discrete embodied pretraining) for the same number of steps (50k for multi-frame modeling), as shown in Fig.\ref{fig:pretraining_importance}. In this comparison, \textit{Scratch (oxe)} refers to training the plain VLM from scratch with multi-frame modeling, on the same OXE datasets as OpenVLA. \textit{Scratch (subset)} denotes training the VLM with multi-frame modeling on the Fractal and Bridge-v2 datasets. The average SimplerEnv score of our method after training for 50k steps is 70.9, while models trained from scratch fail to converge within the limited training budget (10.4 v.s. 70.9 and 18.1 v.s. 70.9). We attribute this to the absence of prior embodied knowledge for training from scratch. Forcing the plain VLM to simultaneously learn embodied perception and multi-frame representations is an inherently challenging process within a limited training budget. Notably, even single-frame models, such as OpenVLA~\cite{openvla}, typically require 28 to 40 epochs to pretrain on OXE, whereas our setting with 50k steps amounts to fewer than 5 epochs, making training from scratch particularly challenging. Our method can leverage pretrained off-the-shelf VLA checkpoints (e.g.,~\cite{openvla,spatialvla,belkhale2024minivla}) in the community, making it a more practical and resource-efficient alternative.

\noindent\textbf{Pretraining without multi-frame regularization.} We also investigate the impact of our multi-frame regularization on training a multi-frame VLA entirely from scratch. Without this regularization, as shown by the SimplerEnv scores of \textit{Scratch (subset) w/o reg.} and \textit{Scratch (oxe) w/o reg.} in Fig.\ref{fig:pretraining_importance}, both settings exhibit lower average success rates and a tendency toward slower convergence compared to their regularized counterparts. These results further highlight the effectiveness of multi-frame regularization in facilitating stable and efficient multi-frame training.

\vspace{-15pt}

\subsection{Ablation on Types of Learnable Feature} Building on the CronusVLA 0.5B model with a minimal frame count (1 past frame, a total of 2), we investigate the impact of different learnable feature representations on performance, focusing on settings with limited temporal context. Specifically, we compare three configurations: (1) \textit{final}, which uses the last-layer of a single learnable feature placed after all visual and language tokens; (2) \textit{visual + final}, which uses two learnable features, and one inserted after the final visual token and the other after all tokens, all features are extracted from the last layer; and (3) \textit{multi-layer final}, which extracts the same position’s features from three different LLM layers (layers 24, 16, and 8 out of 24). As shown in Tab.\ref{tab:ablation_motion_feature}, \textit{multi-layer final} improves performance but increases model complexity, reducing overall usability. Additionally, incorporating extra visual tokens degrades performance, possibly due to misalignment with the VLM’s pretrained causal structure, increasing optimization difficulty. Balancing performance and efficiency, we adopt the \textit{final} strategy in our main model.

\begin{table}[h]
\centering
\renewcommand{\arraystretch}{1.2}
\setlength{\tabcolsep}{4pt}
\vspace{2mm}
\resizebox{0.6\linewidth}{!}{
\begin{tabular}{ccccc}
\toprule
Representation & Google VM & Google VA & WidowX VM & Avearge \\ \midrule
visual + final & 51.3 & 42.4 & 28.1 & 40.6\\
multi-layer final & 65.3 & 60.3 & 39.6 & 55.1\\ \bottomrule \rowcolor{gray!15}
final & 61.4 & 56.6 & 40.6 & 52.9\\ \bottomrule
\end{tabular}
}
\caption{\textbf{Ablation on learnable feature types.}}
\label{tab:ablation_motion_feature}
\end{table}

\subsection{Ablation on Post-training Data}
As shown in Tab.\ref{tab:ablation_training_data}, we evaluate the performance of our model in SimplerEnv under different post-training datasets. All experiments are conducted using the CronusVLA 0.5B model with a total two-frame input. We consider three configurations: (1) training on Bridge-v2~\cite{bridgev2} and evaluating on the WidowX robot, (2) training on Fractal~\cite{RT1} and evaluating on the Google robot, and (3) joint training on both datasets with evaluation on both platforms. Results indicate that joint training consistently improves performance across robots. We attribute this to the complementary nature of the datasets, which facilitates learning more robust and generalized representations, thereby mitigating overfitting associated with single-dataset training.

\begin{table}[h]
\renewcommand{\arraystretch}{1.2}
\setlength{\tabcolsep}{4pt}
\vspace{2mm}
\centering
\begin{tabular}{ccccc}
\toprule
\makecell[c]{Fractal} & \makecell[c]{Bridge} & Google VM & Google VA & WidowX VM \\ \midrule
\cmark &  & 57.8 & 52.0 & - \\
 & \cmark & - & - & 33.3 \\ \bottomrule \rowcolor{gray!15}
\cmark & \cmark & 61.4 & 56.6 & 40.6\\ \bottomrule
\end{tabular}
\caption{\textbf{Ablation on post-training data.}}
\label{tab:ablation_training_data}
\end{table}

\subsection{Detailed Main Result in SimplerEnv} 
\begin{figure*}[t]
    \centering
    \includegraphics[width=1\linewidth]{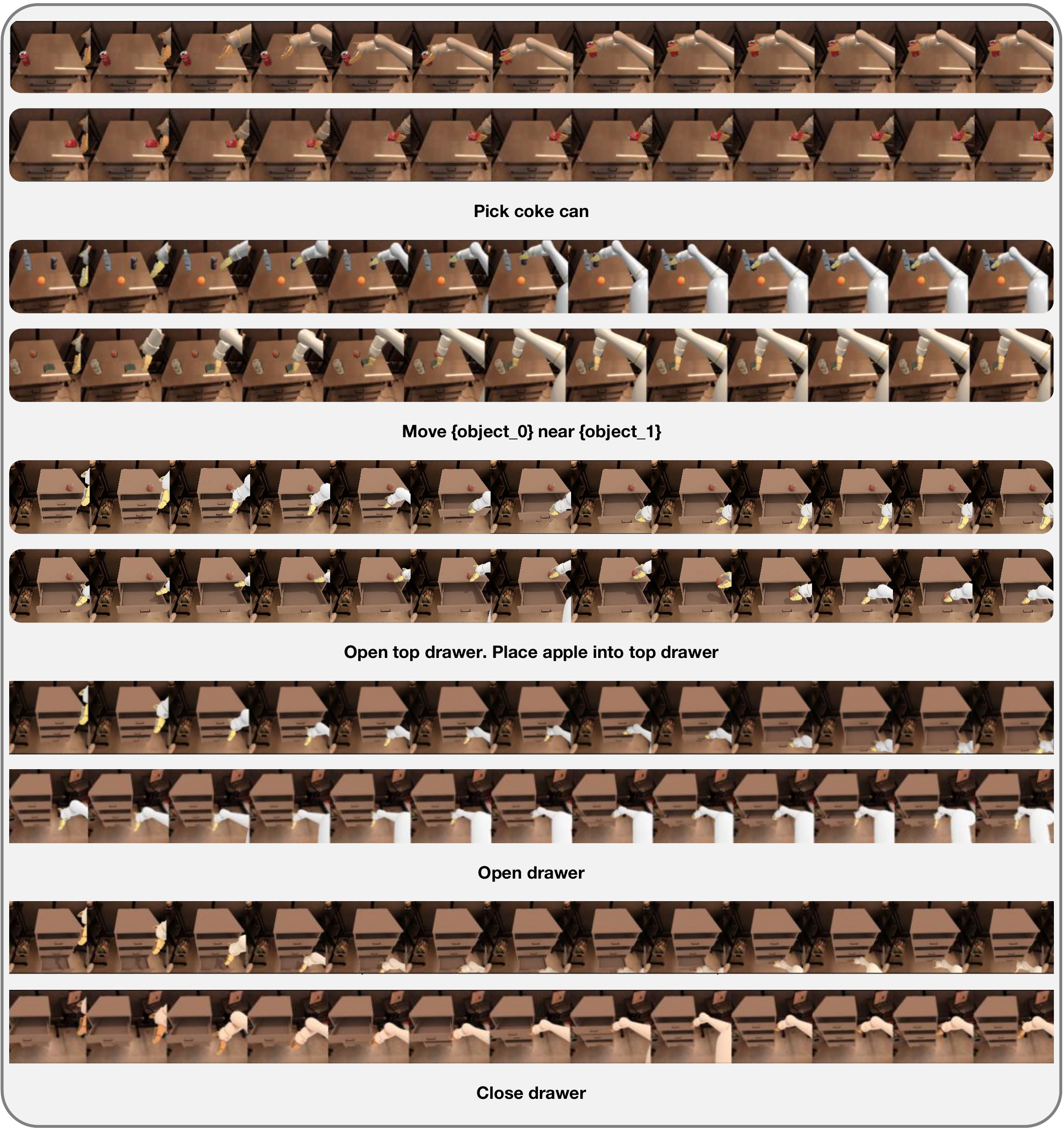}
    \caption{\textbf{Visualizations of Google Robot Visual Matching tasks in SimplerEnv.}}
    \label{fig:visual_GR_VM}
\end{figure*}

\begin{figure*}[t]
    \centering
    \includegraphics[width=1\linewidth]{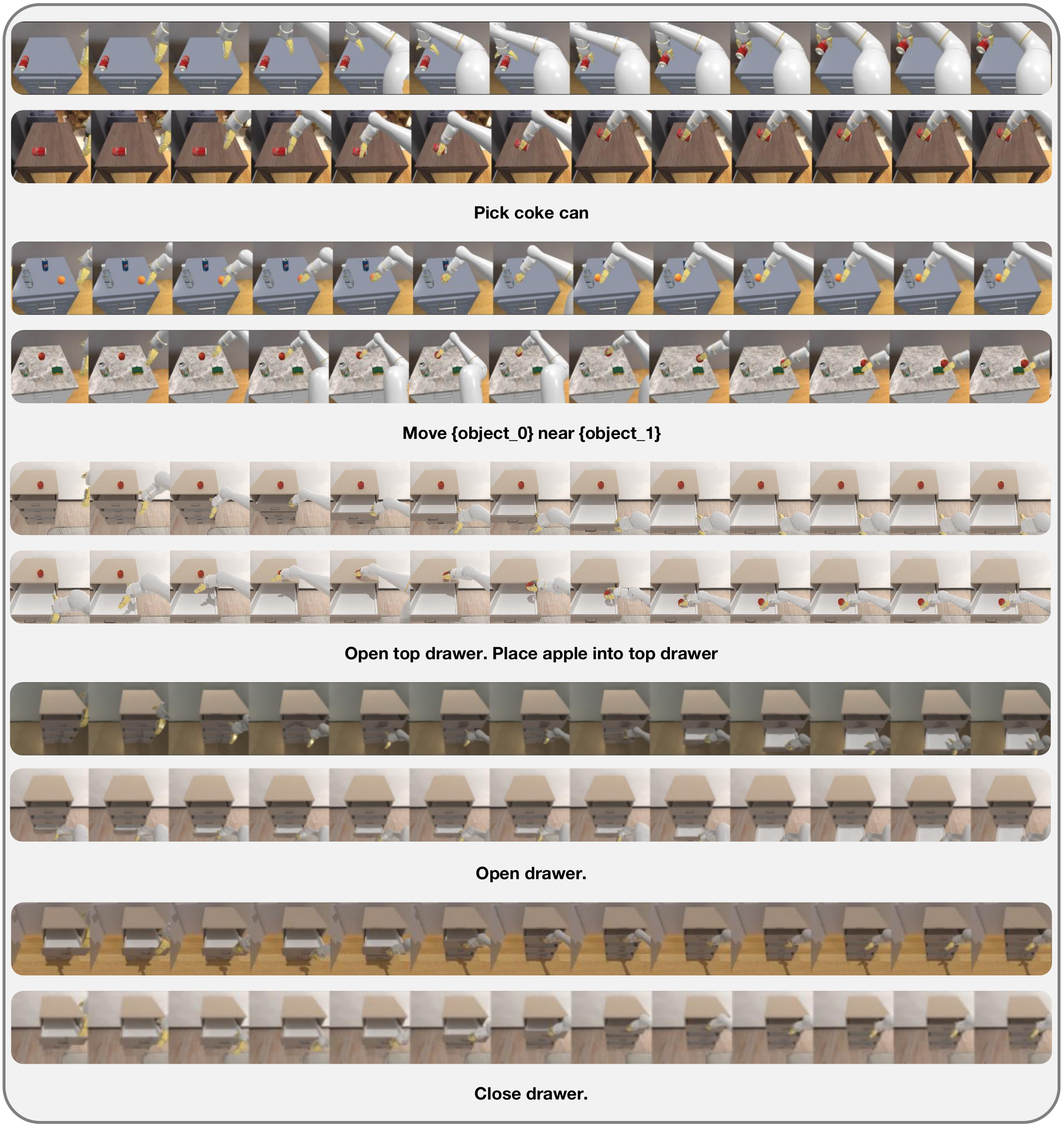}
    \caption{\textbf{Visualizations of Google Robot Variant Aggregation tasks in SimplerEnv.}}
    \label{fig:visual_GR_VA}
\end{figure*}

\begin{figure*}[t]
    \centering
    \includegraphics[width=1\linewidth]{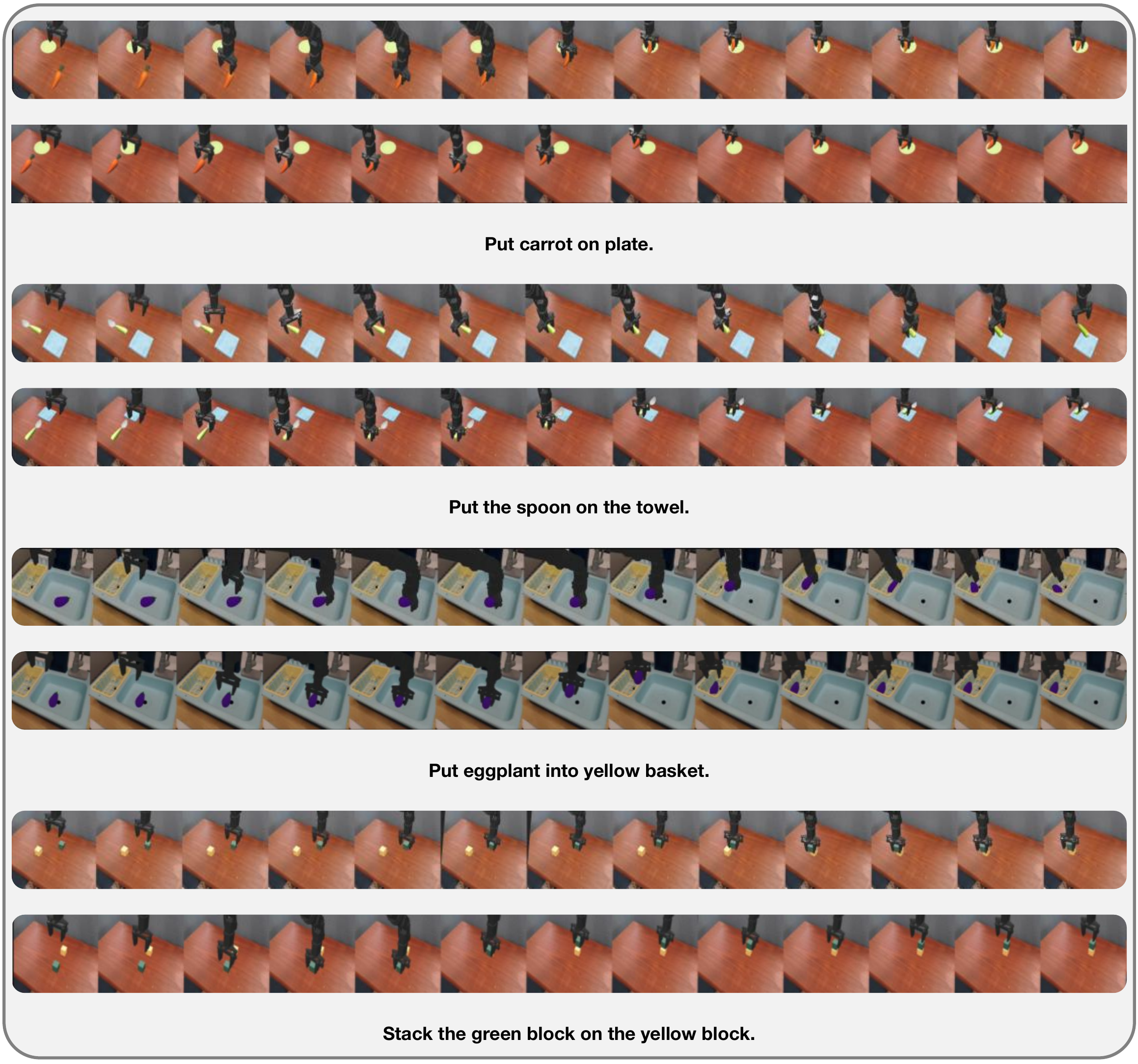}
    \caption{\textbf{Visualizations of WidowX Robot Visual Matching tasks in SimplerEnv.}}
    \label{fig:visual_WR_VM}
\end{figure*}

We report detailed evaluation results for all SimplerEnv settings, focusing on key baselines for comparison. Tab.\ref{tab:detailed_results_GR} presents results on the Google Robot across both Visual Matching and Variant Aggregation, covering four tasks. Extended evaluations include coke can manipulation (horizontal, vertical, and upright grasping) and drawer manipulation (opening and closing). Visualization examples are provided in Fig.\ref{fig:visual_GR_VM} and Fig.\ref{fig:visual_GR_VA}. Additionally, Tab.\ref{tab:detailed_results_WR} reports results on the WidowX Robot, comparing our models against representative baselines. More visualization results are shown in Fig.\ref{fig:visual_WR_VM}. Metrics include grasp success rate and overall task success. Averaged across all tasks, both CronusVLA 0.5B and CronusVLA 7B outperform prior baselines in most cases, demonstrating the effectiveness of our training strategy and model architecture in achieving robust performance and generalization.

\begin{table}[h]
\centering
\renewcommand{\arraystretch}{1.2}
\setlength{\tabcolsep}{4pt}
\vspace{2mm}
\resizebox{0.99\linewidth}{!}{
\begin{tabular}{c|l|ccccccccccc}
\toprule
\multirow{2}{*}{Settings}
& \multirow{2}{*}{Method} 
& \multicolumn{4}{c}{\textbf{Pick Coke Can}} 
& \textbf{Move Near} 
& \multicolumn{3}{c}{\textbf{Open / Close Drawer}} 
& \textbf{Put in Drawer} 
& \textbf{Overall} \\
\cmidrule(lr){3-6} \cmidrule(lr){7-7} \cmidrule(lr){8-10} \cmidrule(lr){11-11} \cmidrule(lr){12-12}
& & \makecell[c]{Horizontal\\Laying}  & \makecell[c]{Vertical\\Laying} & Standing & Average & Average & Open & Close & Average & Average & Average\\
\midrule

\multirow{9}{*}{\makecell[c]{Visual\\Matching}}& RT-1-X & 82.0 & 33.0 & 55.0 & 56.7 & 31.7 & 29.6 & 89.1 & 59.7 & 21.3 & 42.4 \\
 & RT-2-X & 74.0 & 74.0 & 88.0 & 78.7 & 77.9 & 15.7 & 34.3 & 25.0 & 3.7 & 46.3 \\
 & Octo-Base & 21.0 & 21.0 & 9.0 & 17.0 & 4.2 & 0.9 & 44.4 & 22.7 & 0.0 & 11.0 \\
 & OpenVLA & 29.0 & 8.0 & 40.0 & 25.7 & 55.0 & 53.7 & 65.7& 59.7 & 0.0 & 35.1 \\ 
 & RoboVLMs & 91.0 & 46.0 & 92.0 & 76.3 & 79.0 & 31.5 & 58.3 & 44.9 & 27.8 & 57.0 \\
 & SpatialVLA & 67.0 & 79.0 & 92.0 & 79.3 & 90.0 & 45.4 & 63.9 & 54.6 & 0.0 & 55.6 \\ 
 & Magma & 64.0 & 78.0 & 83.0 & 75.0 & 53.0 & 42.8 & 75.0 & 58.9 & 8.3 & 48.8 \\
 \cmidrule(lr){2-12}\rowcolor{gray!15}
 & CronusVLA (0.5B) & 90.0 & 100.0 & 98.0 & 96.0 & 93.0 & 46.3 & 54.6 & 50.5 & 42.6 & 70.5 \\ \rowcolor{gray!15}
 & CronusVLA (7B) & 95.0 & 94.0 & 98.0 & 95.7 & 76.0 & 62.0 & 93.5 & 77.8 & 64.8 & 78.6 \\
\midrule
\multirow{9}{*}{\makecell[c]{Variant\\Aggregation}}& RT-1-X & 56.9 & 20.4 & 69.8 & 49.0 & 32.3 & 6.9 & 51.9 & 29.4 & 10.1 & 30.2 \\
 & RT-2-X & 82.2 & 75.4 & 89.3 & 82.3 & 79.2 & 33.3 & 37.2 & 35.3 & 20.6 & 54.4 \\
 & Octo-Base & 0.5 & 0.0 & 1.3 & 0.6 & 3.1 & 0.0 & 2.1 & 1.1 & 0.0 & 1.2 \\
 & OpenVLA & 56.9 & 38.2 & 67.1 & 54.1 & 63.0 & 19.0 & 28.0 & 23.5 & 2.9 & 35.9 \\
 & RoboVLMs & 77.3 & 31.1 & 43.6 & 50.7 & 62.5 & 4.2 & 16.4 & 10.3 & 0.0 & 30.9 \\
 & SpatialVLA & 89.8 & 71.1 & 75.1 & 78.7 & 83.0 & 21.7 & 56.6 & 39.2 & 6.3 & 51.8 \\
 & Magma & 55.6 & 68.9 & 81.3 & 68.6 & 78.5 & 46.0 & 72.0 & 59.0 & 24.0 & 57.5 \\
 \cmidrule(lr){2-12} \rowcolor{gray!15}
 & CronusVLA (0.5B) & 92.9 & 97.3 & 93.5 & 94.6 & 78.0 & 23.3 & 50.3 & 36.8 & 21.7 & 57.8 \\ \rowcolor{gray!15}
 & CronusVLA (7B) & 96.9 & 96.9 & 88.9 & 94.2 & 77.0 & 29.6 & 87.8 & 58.7 & 65.1 & 73.8 \\
\bottomrule
\end{tabular}
}
\caption{\textbf{Detailed results of different methods on Google Robot setting of SimplerEnv.} Values are success rates (\%).}
\label{tab:detailed_results_GR}
\end{table}

\begin{table}[h]
\centering
\renewcommand{\arraystretch}{1.2}
\setlength{\tabcolsep}{4pt}
\vspace{2mm}
\resizebox{0.9\linewidth}{!}{
\begin{tabular}{l|cccccccccc}
\toprule
\multirow{2}{*}{Method} 
& \multicolumn{2}{c}{Put Spoon on Towel} 
& \multicolumn{2}{c}{Put Carrot on Plate} 
& \multicolumn{2}{c}{Stack Green on Yellow Block} 
& \multicolumn{2}{c}{Put Eggplant in Basket} 
& \multirow{2}{*}{Average} \\
\cmidrule(lr){2-3} \cmidrule(lr){4-5} \cmidrule(lr){6-7} \cmidrule(lr){8-9}
& Grasp & Success & Grasp & Success & Grasp & Success & Grasp & Success & \\
\midrule
RT-1-X             & 16.7 & 0.0  & 20.8 & 4.2  & 8.3  & 0.0  & 0.0  & 0.0  & 1.1  \\
OpenVLA           & 12.5   & 8.3 & 29.2   & 4.2 & 16.7   & 0.0 & 20.8   & 0.0 & 3.1 \\ 
RoboVLMs           & 70.8 & 50.0 & 45.8 & 37.5 & 70.8 & 0.0  & 91.7 & 83.3 & 42.7 \\
Magma              & 62.5 & 37.5 & 41.7 & 29.2 & 79.2 & 20.8 & 95.8 & 91.7 & 44.8 \\
SpatialVLA         & 25.0 & 20.8 & 41.7 & 37.5 & 79.2 & 41.7 & 91.7& 83.3& 45.8 \\
\bottomrule \rowcolor{gray!15}
CronusVLA (0.5B)   & 54.2 & 45.8 & 54.2 & 33.3 & 58.3 & 0.0  & 79.2 & 79.2 & 39.6 \\
\rowcolor{gray!15}
CronusVLA (7B)     & 75.0 & 66.7 & 79.2 & 54.2 & 41.7 & 20.8 & 100.0& 100.0& 60.4 \\
\bottomrule
\end{tabular}
}
\caption{\textbf{Detailed results of different methods on WidowX Robot setting of SimplerEnv.} Values are success rates (\%).}
\label{tab:detailed_results_WR}
\end{table}

\subsection{More Results in LIBERO Benchmark} 

\begin{table}[h]
  \centering
  \resizebox{0.9\linewidth}{!}{
    \begin{tabular}{c|cccc|c}
      \toprule
      Methods & Spatial & Object & Goal & Long (10) & Ave. \\ \midrule
      & \multicolumn{5}{c}{\textit{Type I: without state, without wrist view}} \\ \midrule 
      Diffusion Policy~\cite{diffusionpolicy} & 78.5 $\pm$ 1.1 & 87.5 $\pm$ 0.7 & 73.5 $\pm$ 1.2 & 64.8 $\pm$ 1.3 & 76.1 $\pm$ 0.7 \\
      OpenVLA~\cite{openvla} & 84.7 $\pm$ 0.9 & 88.4 $\pm$ 0.8 & 79.2 $\pm$ 1.0 & 53.7 $\pm$ 1.3 & 76.5 $\pm$ 0.6 \\
      TraceVLA~\cite{Tracevla} & 84.6 $\pm$ 0.2 & 85.2 $\pm$ 0.4 & 75.1 $\pm$ 0.3 & 54.1 $\pm$ 1.0 & 74.8 $\pm$ 0.5 \\
      SpatialVLA~\cite{spatialvla} & 88.2 $\pm$ 0.5 & 89.9 $\pm$ 0.7 & 78.6 $\pm$ 0.6 & 55.5 $\pm$ 1.0 & 78.1 $\pm$ 0.7 \\
      Dita~\cite{hou2025dita} & 84.2 & 96.3 & 85.4 & 63.8 & 82.4 \\ \midrule\rowcolor{gray!15}
      CronusVLA & 93.8 $\pm$ 0.9 & 92.8 $\pm$ 0.0 & 95.6 $\pm$ 0.4 & 86.5 $\pm$ 0.8 & 92.2 $\pm$ 0.4 \\\midrule
      &\multicolumn{5}{c}{\textit{Type II: with state, with wrist view}} \\ \midrule 
      GR00T-N1~\cite{bjorck2025gr00t} & 94.4 $\pm$ 0.9 & 97.6 $\pm$ 1.0 & 93.0 $\pm$ 1.2 & 90.6 $\pm$ 1.0 & 93.9 $\pm$ 1.1 \\
      $\pi_0$~\cite{pi0} & 96.8 $\pm$ 0.8 & 98.8 $\pm$ 0.9 & 95.8 $\pm$ 1.1 & 85.2 $\pm$ 1.2 & 94.2 $\pm$ 0.9 \\
      $\pi_0$-FAST~\cite{pertsch2025fast} & 96.4 $\pm$ 0.7 & 96.8 $\pm$ 0.7 & 88.6 $\pm$ 1.0 & 60.2 $\pm$ 1.4 & 85.5 $\pm$ 1.0 \\
      OpenVLA-OFT~\cite{openvla-oft} & 97.6 $\pm$ 0.9 & 98.4 $\pm$ 0.8 & 97.9 $\pm$ 1.0 & 94.5 $\pm$ 1.3 & 97.1 $\pm$ 0.6 \\
      $\pi_{0.5}+KI$\ (from scratch)~\cite{intelligence2025pi_} & 96.6 & 97.2 & 94.6 & 84.8 & 93.3 \\
      $\pi_{0.5}+KI$\ (from generalist model)~\cite{intelligence2025pi_} & 98.0 & 97.8 & 95.6 & 85.8 & 94.3 \\ \midrule \midrule\rowcolor{gray!15}
      CronusVLA (only with wrist view) & 97.3 $\pm$ 0.2 & 99.6 $\pm$ 0.0 & 96.9 $\pm$ 0.3 & 94.0 $\pm$ 1.0 & 97.0 $\pm$ 0.4 \\
      \bottomrule
    \end{tabular}
  }
  \caption{\textbf{Detailed LIBERO Benchmark Results.} We report the success rate (SR) and standard error for each method across four task suites, averaged over three random seeds with 500 trials. To ensure fair comparison under different settings, we evaluate two categories of VLA models: Type I, which relies solely on the instruction and third-person view (e.g., OpenVLA, SpatialVLA); and Type II, which additionally incorporates state and wrist-view inputs (e.g., $\pi_0$, OpenVLA-OFT).}
  \label{tab:detailed_libero_results}
\end{table}

\begin{figure*}[t]
    \centering
    \includegraphics[width=1\linewidth]{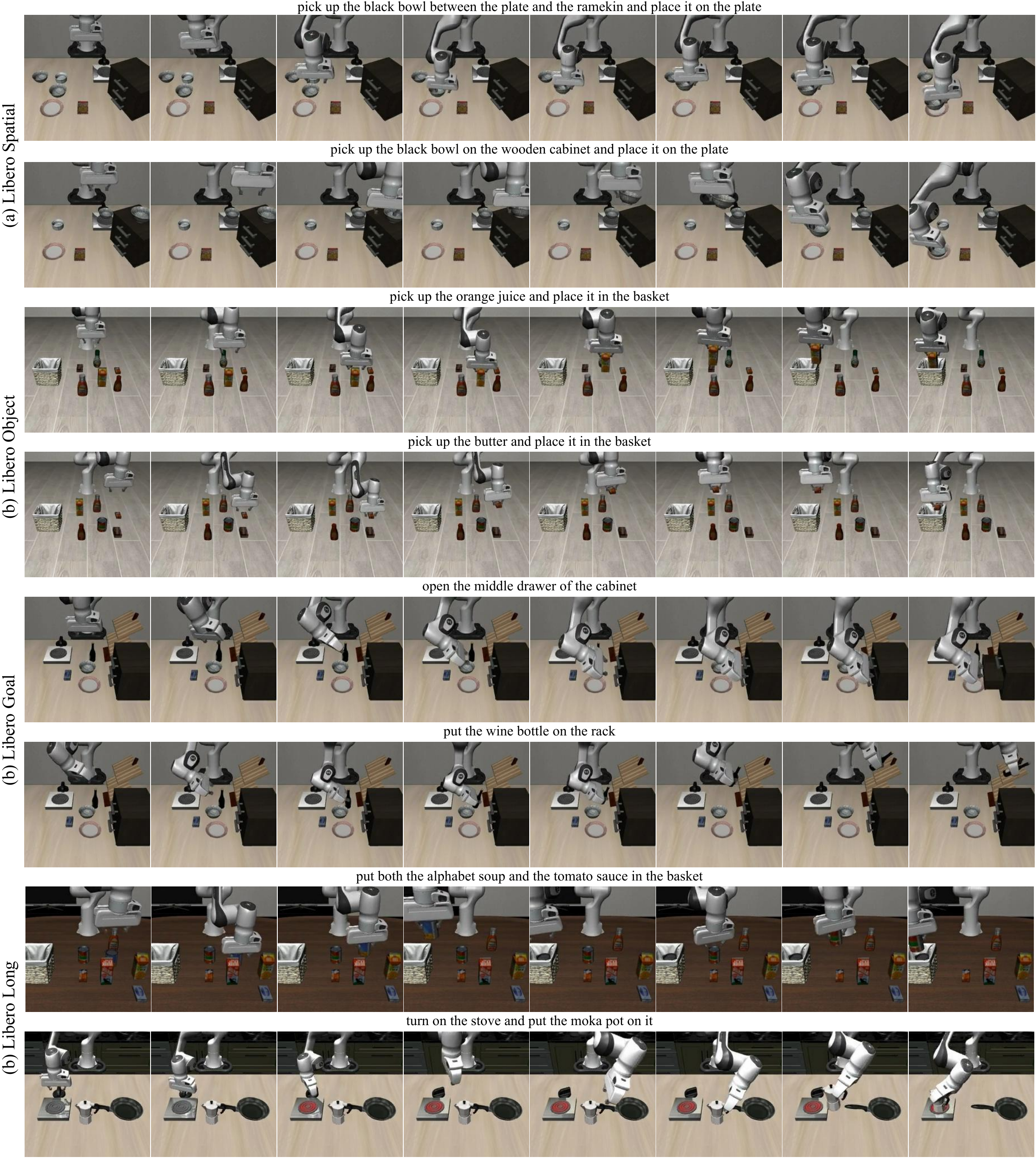}
    \caption{\textbf{Visualizations of Franka Robot, all four task suites in LIBERO.}}
    \label{fig:visual_LIBERO}
\end{figure*}

Detailed results on LIBERO are listed in Tab.\ref{tab:detailed_libero_results}. Our model achieves performance comparable to the prior SOTA OpenVLA-OFT, despite not using state inputs or additional optimization, and requiring substantially fewer training steps (30K vs. 150K). We illustrate some visualization results of four task suites in Fig.\ref{fig:visual_LIBERO}, including LIBERO Spatial, LIBERO Object, LIBERO Goal, and LIBERO Long.

\clearpage
\newpage

\section{Training Details}
\label{sec_Training_Details}
\begin{figure*}[h]
    \centering
    \includegraphics[width=\linewidth]{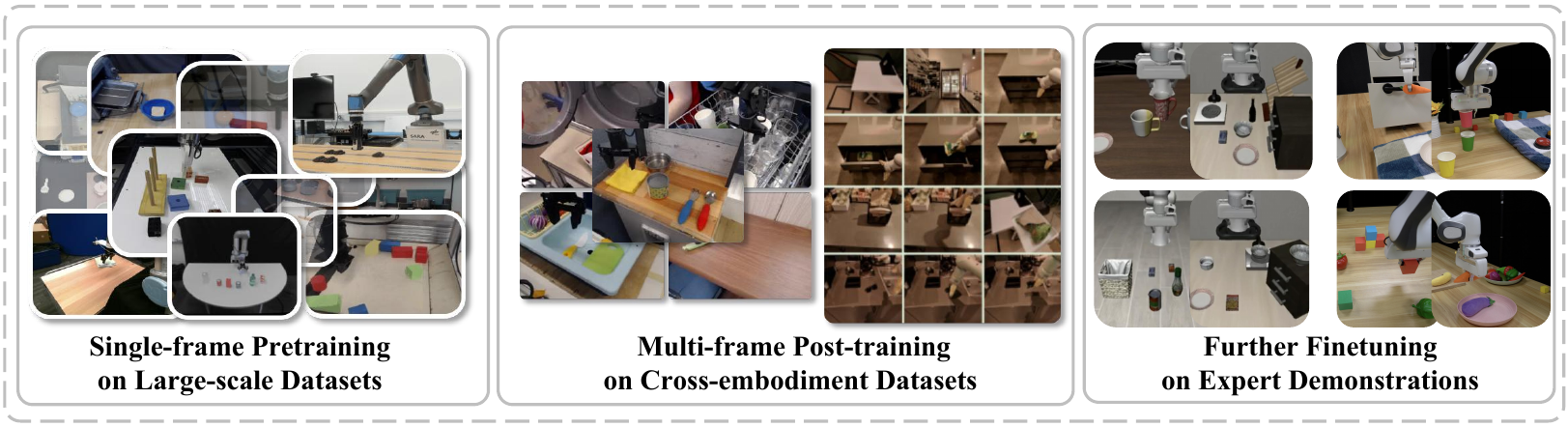}
    \caption{\textbf{Training pipeline.}}
    \label{fig:training_three}
\end{figure*}

The training stages, including single-frame pretraining, multi-frame post-training, and finetuning, are illustrated in the Fig.\ref{fig:training_three}.

\subsection{Pretraining Details} For the 7B model, we follow the training setup and protocol of OpenVLA~\cite{openvla} by leveraging 27 datasets from Open X-Embodiment~\cite{oxe}. The same procedure is applied to the 0.5B model based on the VLM of~\cite{belkhale2024minivla}. Training is performed by minimizing the cross-entropy loss between the predicted token distribution and ground truth tokens. For more details, please refer to~\cite{openvla}.

\subsection{Post-training Details} Our models are post-trained on cross-embodiment datasets, including Fractal\cite{RT2} and Bridge-v2\cite{bridgev2}. We follow the data processing pipeline of Octo~\cite{octo} and OpenVLA~\cite{openvla}, including RLDS preprocessing and augmentations such as random resized cropping, and adjustments to brightness, contrast, saturation, and hue. To enhance generalization, the data loader randomly samples across datasets, trajectories, and time steps. All components, including the vision encoder, projection layer, language model, and decoder, are trained end-to-end by minimizing the mean squared error between predicted and ground-truth diffusion noise. Training employs a 100-step diffusion noise schedule. The language model does not generate text tokens autoregressively; we omit text prediction cross-entropy loss and discard the language head.

CronusVLA 7B is initialized from the pretrained single-frame 7B VLA model. Post-training is conducted on 64 A100 GPUs for approximately 50k gradient steps, with a total batch size of 512 and default 6-frame past observation sequences. We use AdamW with a learning rate of 4e-5 and a linear scheduler, without weight decay or warm-up. PyTorch FSDP is applied to reduce memory usage. Due to the inherent instability of diffusion-based training, which is sensitive to randomness, we select the best-performing checkpoint between 45k and 55k steps, evaluated every 2.5k steps. CronusVLA 0.5B utilizes Qwen2.5 0.5B~\cite{qwen2} as the LLM backbone, SigLIP~\cite{siglip} and DINOv2~\cite{dinov2} as the vision encoder. Post-training is performed on 32 A100 GPUs, about 40k gradient steps, with a batch size of 1024 and default 3-frame past observation sequences.  Action chunking~\cite{ACT} is employed by following the settings of \cite{cogact} with a chunking length of 16 and an adaptive ensemble strategy. We also use AdamW with a learning rate of 4e-5 and a linear scheduler, without weight decay or warm-up.

\subsection{Finetuning Details} Building upon the post-trained models, we further finetune on expert demonstrations using a full finetuning strategy. The post-trained weights are adapted to the specific task setup (such as each task suite of LIBERO). CronusVLA-7B is finetuned for 6K-30K steps. Training is conducted with a batch size of 256 on 16 A100 GPUs, using a constant learning rate of 2e-5 and a fixed sequence length of 3-frame past observations, and the action chunking length is 8.

\clearpage
\newpage

\section{Model Details}
\label{sec_Model_Details}

\subsection{Model Architecture} As shown in Tab.\ref{tab:model_arch}, we detail the model architecture. The language model is either LLaMA2 or Qwen2.5, and the vision encoder combines SigLIP and DINOv2, taking 224×224 RGB images as input. A linear layer serves as the cross-modal projector (MM projector). These components comprise the core VLA module, which processes $B \cdot F$ individual samples. The decoder integrates multiple inputs: ground-truth actions, timestep encodings, and multi-frame learnable features from the VLA. Actions and timesteps are embedded into a 768-dimensional space via the Action and Timestep Embedders, respectively. The Action Adapter is activated only during finetuning. Learnable features are modulated by the Modulator, and the decoder consists of 12 layers of Self-Attention and Cross-Attention, facilitating interaction between noised actions and learnable features. MLP in our model includes two linear layers with the GELU function. Modulator includes a DIV function(as shown in Eq.\ref{eque_2}) and a learnable positional embedding (as shown in Fig.\ref{fig:method} (d)).

\renewcommand{\arraystretch}{1.2}
\begin{table}[ht]
\centering
\small
\vspace{2mm}
\begin{tabular}{@{}p{3.2cm}p{2.4cm}p{1.7cm}p{2.3cm}p{2.3cm}@{}}
\toprule
\textbf{Module (Parameters)} & \textbf{Submodule} & \textbf{Layer Num} & \textbf{Input Shape} & \textbf{Output Shape} \\
\midrule

\multirow{2}{*}{Vision Encoder (731M)} 
  & SigLIP  & 1+27+1 &  \multirow{2}{*}{$(B\cdot F,3,H,W)$}&  \multirow{2}{*}{$(B\cdot F,256,2176)$}\\
  & Dinov2  & 1+24+1 \\
\midrule
MM Projector (71M)
  & Linear (for 7B) & \multirow{2}{*}{1}  
  & \multirow{2}{*}{$(B\cdot F,256,2176)$} 
  & $(B\cdot F,256,4096)$ \\
MM Projector (28M)
  & Linear (for 0.5B) & 
  &
  & $(B\cdot F,256,896)$ \\
\midrule
LLM Backbone (6738M)
  & LLaMA2   & 1+32+0 
  & \multirow{2}{*}{$(B\cdot F,256+T)$} 
  & $(B,F,4096)$ \\
LLM Backbone (630M)
  & Qwen2.5  & 1+24+0 
  &
  & $(B,F,896)$ \\
\midrule
\multirow{7}{*}{Decoder (135M)}
  & Action Embed.     & 1  & $(B,A,7)$               & $(B,A,768)$ \\
  & Timestep Embed.   & 1  & $(B,256)$                 & $(B,1,768)$ \\
  & Action Adapter      & 1  & $(B,A,7+7)$              & $(B,A,768)$ \\
  & Modulator           & 1  & $(B,F,4096\text{ or }896)$  & $(B,2F,768)$ \\
  & Self-Attn. + MLP     & 12 & $(B,A+1,768)$           & $(B,A+1,768)$ \\
  & Cross-Attn.     & 12 & $(B,A+1,768)+(B,2F,768)$ & $(B,A+1,768)$ \\
  & Final Layer         & 1  & $(B,A+1,768)$           & $(B,A+1,7)$ \\
\bottomrule
\end{tabular}
\caption{\textbf{Model architecture components} with model size, layer numbers, and tensor shapes. Notation: $B$ = batch size, $F$ = total number of past and current frames, $T$ = text token count, $A$ = action sequence length.}
\label{tab:model_arch}
\end{table}

\subsection{Feature Chunking} 
\begin{figure*}[h]
    \centering
    \includegraphics[width=0.3\linewidth]{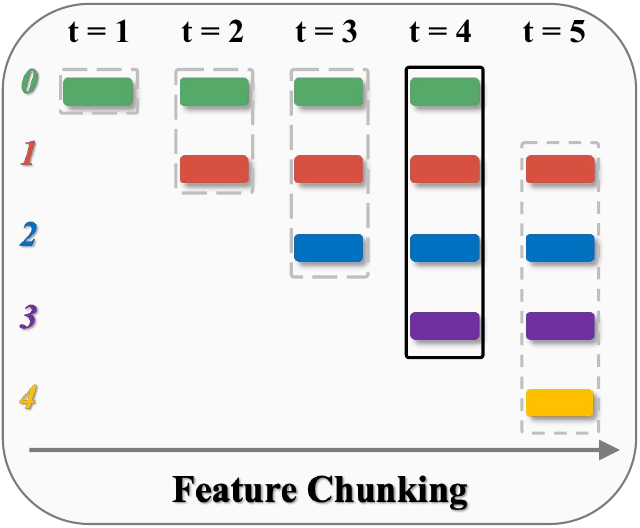}
    \caption{\textbf{Details of the feature chunking mechanism.} During inference, the model processes a sequence consisting of several previous frames and one current frame.}
    \label{fig:model_details}
\end{figure*}

The core logic of feature chunking is depicted in Fig.\ref{fig:model_details} (b). At inference, feature chunking accelerates prediction by caching previously computed learnable features. To maintain a consistent feature length M at each step, a first-in-first-out queue is employed. For early timesteps ($t<M$) with insufficient historical features, we adopt a standard low-level policy strategy by padding the buffer with repeated first-frame features. As similar padding is introduced during training via stochastic sampling, this method does not degrade performance.

\subsection{Efficiency Analysis}

\noindent
Our model improves both performance and inference efficiency, and supports limited-cost training. An additional comparison is shown in Table~\ref{tab:efficiency}. 
Our single-frame pretraining can be replaced by reusing or sharing public checkpoints (e.g., \textit{OpenVLA}~\cite{openvla}), while the post-training stage requires only $\sim$50K steps, which is significantly fewer than other VLA counterparts. 
Our design supports not only a compact 0.5B model but also a larger 7B model. As the history length of CronusVLA-7B increases from 2 to 7, the inference speed remains constant (8.7~Hz) in Fig.\ref{fig:ablation_study}. 
The proposed \textit{multi-frame regularization} employs \texttt{no\_grad} for past frames with only one forward pass, ensuring efficient and scalable training.

\begin{table}[h!]
\centering
\small
\setlength{\tabcolsep}{4.5pt}
\begin{tabular}{lccccccc}
\toprule
\textbf{Method} & \makecell{\textbf{Frame}\\ \textbf{Number}} & \makecell{\textbf{Success}\\ \textbf{Rate$\uparrow$}} & \makecell{\textbf{Inference}\\ \textbf{Speed (Hz)$\uparrow$}}&  \makecell{\textbf{Inference}\\ \textbf{Latency (ms)$\downarrow$}}& \makecell{\textbf{Inference}\\ \textbf{Memory (GB)$\downarrow$}} & \makecell{\textbf{Training}\\ \textbf{Step (K)}} \\
\midrule
OpenVLA & 1 & 24.7 & 5.2& 192.3& 15.0 & 295 \\
+MultiFrame & 7 & 32.4 & 3.1& 263.2 & 17.2 & 50\\
SpatialVLA & 1 & 51.2 & 2.4& 322.5 & 8.6 & 200\\
Ours-0.5B & 4 & 56.0 & 11.1& 90.1 & 3.7 & 40\\
Ours-7B & 7 & \textbf{70.9} & 8.7& 114.9 & 15.1 & 50\\
\bottomrule
\end{tabular}
\caption{
\textbf{Comparison of performance, efficiency, and cost.} Inference is conducted on A100 GPUs without chunking, and success rates on SimplerEnv are reported. Training steps are summarized from official setups. Different VLA models with varying frame numbers and latency–accuracy trade-off are also listed.
}
\label{tab:efficiency}
\end{table}

\clearpage
\newpage

\section{Details of Real-world Experiments}
\label{sec_Details_of_Real_world_Experiments}
This section details the evaluation setup for real-world experiments on the Franka platform.

\subsection{Experimental Setup}

\begin{figure*}[h]
    \centering
    \includegraphics[width=0.6\linewidth]{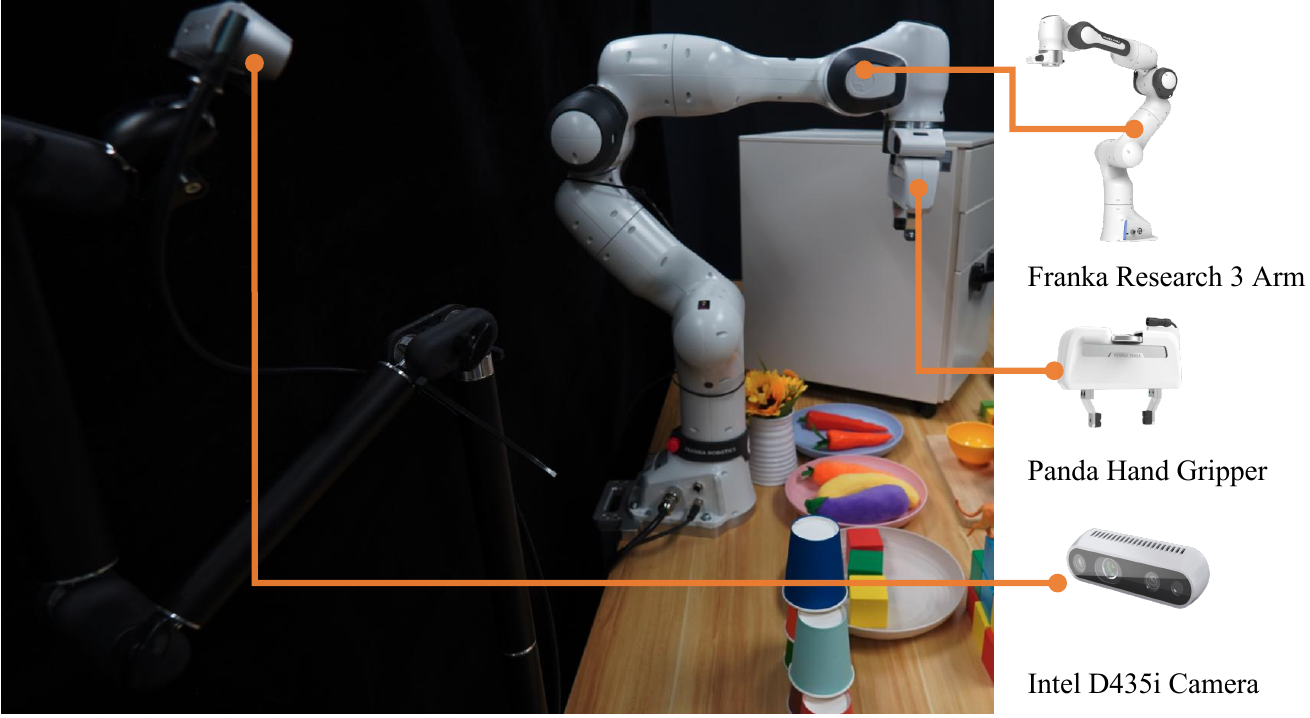}
    \caption{\textbf{Our Franka platform.}}
    \label{fig:franka_platform}
\end{figure*}

\noindent\textbf{Model deployment.} We evaluate our method on several real-world tasks using the Franka Research 3 Robot equipped with a 7-DoF arm and a 1-DoF Panda Hand gripper. Visual input is provided by a single Intel D435i camera configured in an eye-on-base setup, as shown in Fig.\ref{fig:franka_platform}. Both OpenVLA~\cite{openvla} and CronusVLA require over 10 GB of memory in bfloat16 precision; thus, all models are deployed on an online A100 GPU server, while a local machine handles robot control. Although our method and DP3 support higher frequencies, we limit both data collection and action execution to 5 Hz to match OpenVLA’s speed constraints and mitigate communication latency. Finetuning data are collected via human teleoperation using a SpaceMouse device.

\noindent\textbf{Training details.} For CronusVLA, we follow the finetuning protocol from the LIBERO experiments, initializing CronusVLA-7B with post-trained weights and conducting cross-embodiment finetuning on our collected demonstrations. For DP3\cite{dp3}, we adopt the updated instruction-conditioned architecture from\cite{ke20243d} and implement it ourselves. We report the best-performing checkpoint of DP3. For OpenVLA~\cite{openvla}, we follow the official training protocol, fully finetuning the model until the token accuracy reaches 95\%, and report its best checkpoint. All models are trained using a consistent data augmentation strategy.

\subsection{Tasks Settings}

\begin{figure*}[t]
    \centering
    \includegraphics[width=1\linewidth]{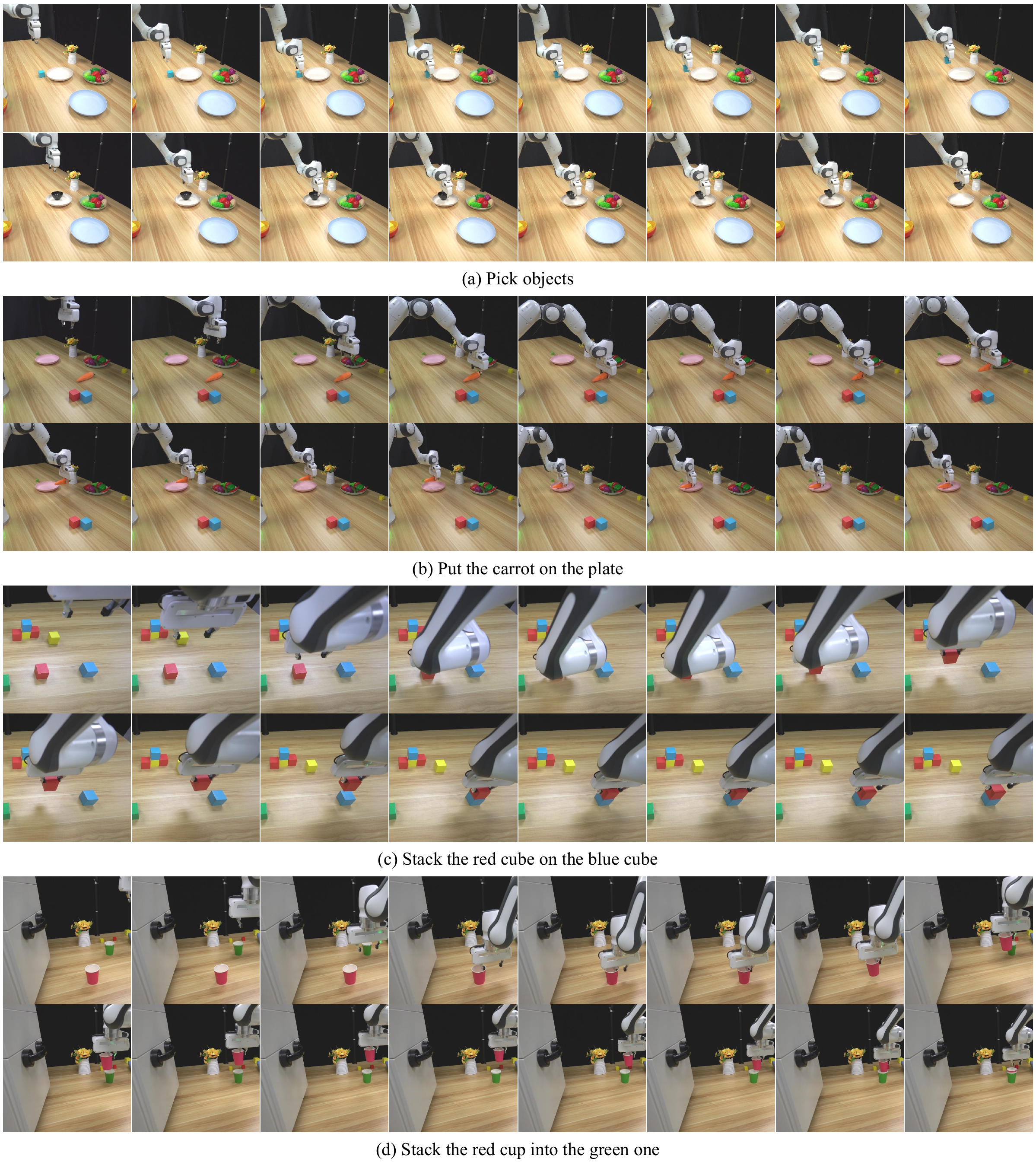}
    \caption{\textbf{Visualizations of real-world Simple Pick-and-Place tasks.}}
    \label{fig:visual_real_world_simple}
\end{figure*}

\noindent\textbf{Simple Pick-and-Place.}
These tasks involve straightforward pick-and-place actions with at most two sub-tasks. Detailed success rates are shown in Tab.\ref{tab:pick_place_results}, visualizations are shown in Fig.\ref{fig:visual_real_world_simple} and our video demo. We evaluate the model’s learning capability and spatial generalization through simple manipulation tasks:

\begin{itemize}

\item \textbf{Pick objects.} This task assesses the model’s fundamental abilities in localization, grasping, and spatial generalization. Four target objects are used: an eggplant, a carrot, a wooden cube, and a bowl. The eggplant and carrot are relatively easy to grasp, while the cube and bowl demand higher placement precision to prevent arm jamming or slippage. Object positions are changed within a region centered on the plate. Illustrated in Fig.\ref{fig:visual_real_world_simple} (a).

\item \textbf{Put the carrot on the plate.} This task primarily evaluates grasping precision and placement accuracy for the single-object pick-and-place. The robot must grasp a carrot from a tabletop region on the right, where objects are positioned with varying locations and orientations, and place it into the designated bowl. Illustrated in Fig.\ref{fig:visual_real_world_simple} (b).

\item \textbf{Stack the red cube on the blue cube.} This task poses a greater challenge than the previous pick-and-place setup. The robot is required to accurately grasp a red cube from a designated area and stack it on a randomly positioned blue cube. The varying positions of both cubes across trials increase task complexity, especially for models relying exclusively on third-person visual observations. Illustrated in Fig.\ref{fig:visual_real_world_simple} (c).

\item \textbf{Stack the red cup into the green one.} This task evaluates the model’s capabilities of instruction following and spatial perception. We introduce considerable spatial variability by randomly placing the red, green, and yellow cups within three predefined regions and frequently permuting their positions. During data collection, either the red or yellow cup is randomly placed into the green cup, with corresponding language instructions annotated. And evaluation is restricted to the instruction, "stack the red cup into the green one". Illustrated in Fig.\ref{fig:visual_real_world_simple} (d).
\end{itemize}

\begin{table}[h]
\centering
\renewcommand{\arraystretch}{1.2}
\setlength{\tabcolsep}{4pt}
\vspace{2mm}
\resizebox{0.6\linewidth}{!}{
\begin{tabular}{l|cccccccccc}
\toprule
\multirow{2}{*}{Method} & \multicolumn{4}{c}{\textbf{Pick Objects}} & \multicolumn{2}{c}{\textbf{Put Carrot}} & \multicolumn{2}{c}{\textbf{Stack Cubes}} & \multicolumn{2}{c}{\textbf{Stack Cups}} \\
\cmidrule(lr){2-5} \cmidrule(lr){6-7} \cmidrule(lr){8-9} \cmidrule(lr){10-11}
& Eggplant & Carrot & Cube & Bowl & Pick & Place & Pick & Place & Pick & Place \\
\midrule
DP3     & 72.0 & 80.0 & 40.0 & 28.0 & 84.0 & 72.0 & 36.0 & 12.0 & 4.0  & 0.0  \\
OpenVLA & 76.0 & 68.0 & 52.0 & 20.0 & 68.0 & 64.0 & 44.0 & 28.0 & 32.0 & 0.0  \\ \bottomrule \rowcolor{gray!15}
Ours    & 72.0 & 84.0 & 64.0 & 40.0 & 80.0 & 80.0 & 56.0 & 48.0 & 28.0 & 16.0 \\
\bottomrule
\end{tabular}
}
\caption{\textbf{Performance comparison across Simple Pick-and-Place tasks.} Values are success rates (\%).}
\label{tab:pick_place_results}
\end{table}

\begin{figure*}[t]
    \centering
    \includegraphics[width=1\linewidth]{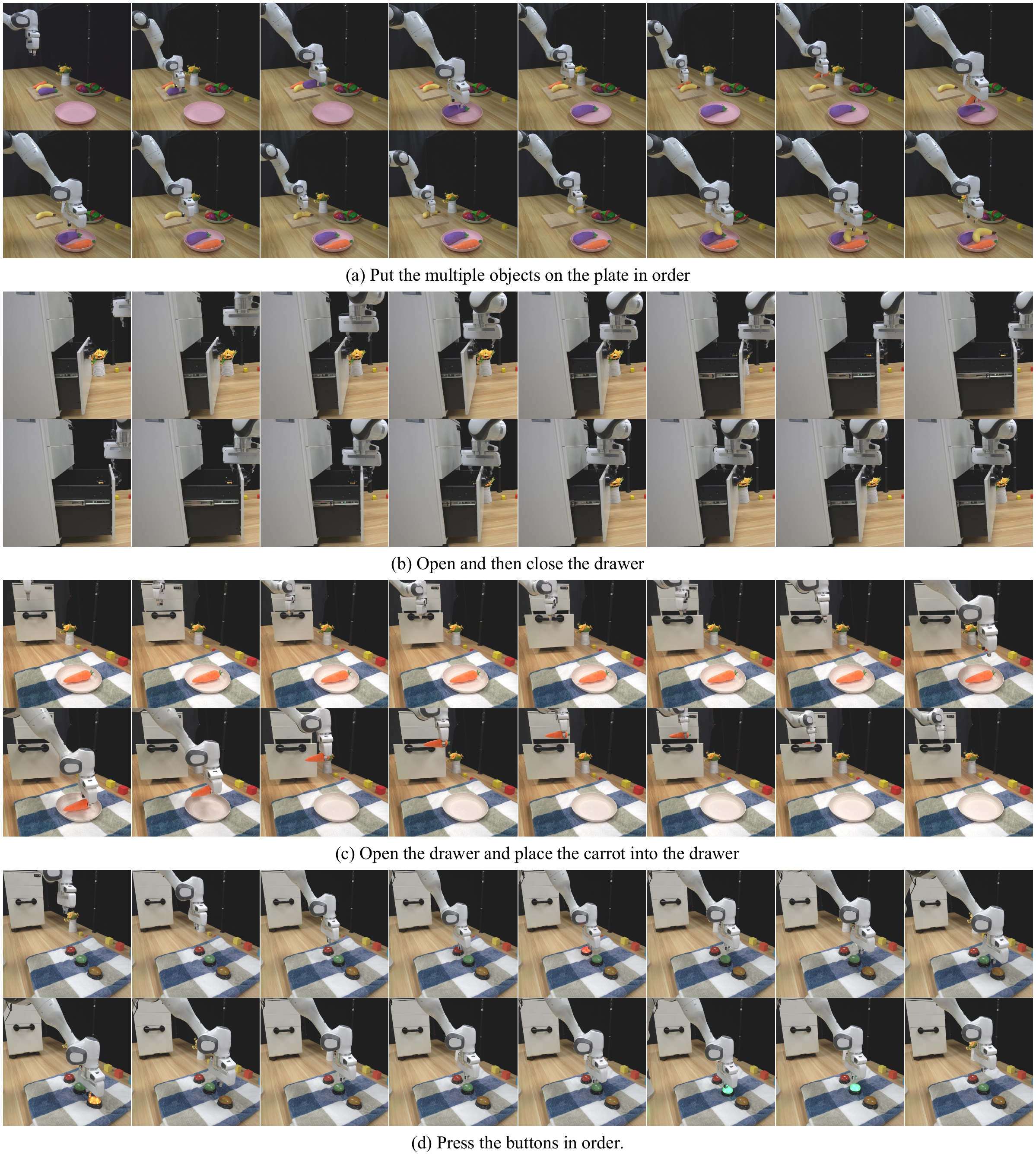}
    \caption{\textbf{Visualizations of real-world Long-horizon tasks.}}
    \label{fig:visual_real_world_long}
\end{figure*}

\noindent\textbf{Long-horizon Tasks.} We evaluate tasks involving multi-step object manipulation with at least two sub-tasks to evaluate the model’s proficiency in sequential execution and long-horizon task composition. Detailed success rates are shown in Tab.\ref{tab:long_horizon_results}, visualizations are shown in Fig.\ref{fig:visual_real_world_long} and our video demo:

\begin{itemize}
\item \textbf{Put the multiple objects on the plate in order.} This task assesses the model’s ability of multi-step execution and instructions following. It extends the single-object pick-and-place setup to a multi-object setting. During data collection, one object is randomly selected for each grasping step until all objects are relocated, with varying relative placements in the bowl. In evaluation, the instruction of grasping order is fixed. Illustrated in Fig.\ref{fig:visual_real_world_long} (a).
\item \textbf{Open and then close the drawer.} This task evaluates the model’s ability to manipulate articulated objects through sequential actions, emphasizing precise localization of interaction points. The robot must grasp a drawer handle, pull to open, release and reposition the gripper, and push to close the drawer. During both training and evaluation, the drawer is rotated differently, with its initial opening set to one of three predefined states. Success is defined by the correct execution of both opening and closing. Data collection is split into two stages, while evaluation follows a unified sequential protocol. Illustrated in Fig.\ref{fig:visual_real_world_long} (b).
\item \textbf{Open the drawer and place the carrot into the drawer.} This task assesses the model’s stability across a wide spatial range and its capability in long-horizon manipulation. The robot must sequentially pull open a drawer, grasp an object from the plate, and place it into the drawer. Variations include the carrot’s position and orientation within the plate, as well as the drawer’s initial opening state. Illustrated in Fig.\ref{fig:visual_real_world_long} (c).
\item \textbf{Press the buttons in order.} This evaluation primarily measures the model’s ability to disambiguate observations and accurately follow instructions. The robot sequentially presses three color-coded buttons whose positions vary within a defined range. The pressing order of testing uses a fixed sequence: red, yellow, then green. The task introduces substantial visual ambiguity, as similar observations occur at different stages, such as before and after button presses. Illustrated in Fig.\ref{fig:visual_real_world_long} (d).
\end{itemize}

\begin{table}[h]
\centering
\renewcommand{\arraystretch}{1.2}
\setlength{\tabcolsep}{4pt}
\vspace{2mm}
\resizebox{0.8\linewidth}{!}{
\begin{tabular}{l|cccccccccc}
\toprule
\multirow{2}{*}{Method} 
& \multicolumn{3}{c}{\textbf{Multiple Pick-Place}} 
& \multicolumn{2}{c}{\textbf{Open-Close Drawer}} 
& \multicolumn{2}{c}{\textbf{Open and Place in Order}} 
& \multicolumn{3}{c}{\textbf{Press Button in Order}} \\
\cmidrule(lr){2-4} \cmidrule(lr){5-6} \cmidrule(lr){7-8} \cmidrule(lr){9-11}
& First & Second & Third & Open & Close & Open & Place & First & Second & Third \\
\midrule
DP3     & 12.0 & 4.0  & 0.0  & 32.0 & 8.0  & 24.0 & 0.0  & 16.0 & 8.0  & 8.0  \\
OpenVLA & 40.0 & 16.0 & 4.0  & 60.0 & 48.0 & 52.0 & 24.0 & 72.0 & 64.0 & 40.0 \\ \bottomrule \rowcolor{gray!15}
Ours    & 52.0 & 28.0 & 20.0 & 72.0 & 64.0 & 72.0 & 60.0 & 96.0 & 92.0 & 88.0 \\
\bottomrule
\end{tabular}
}
\caption{\textbf{Performance comparison across real-world Long-horizon tasks.}}
\label{tab:long_horizon_results}
\end{table}

\begin{figure*}[t]
    \centering
    \includegraphics[width=1\linewidth]{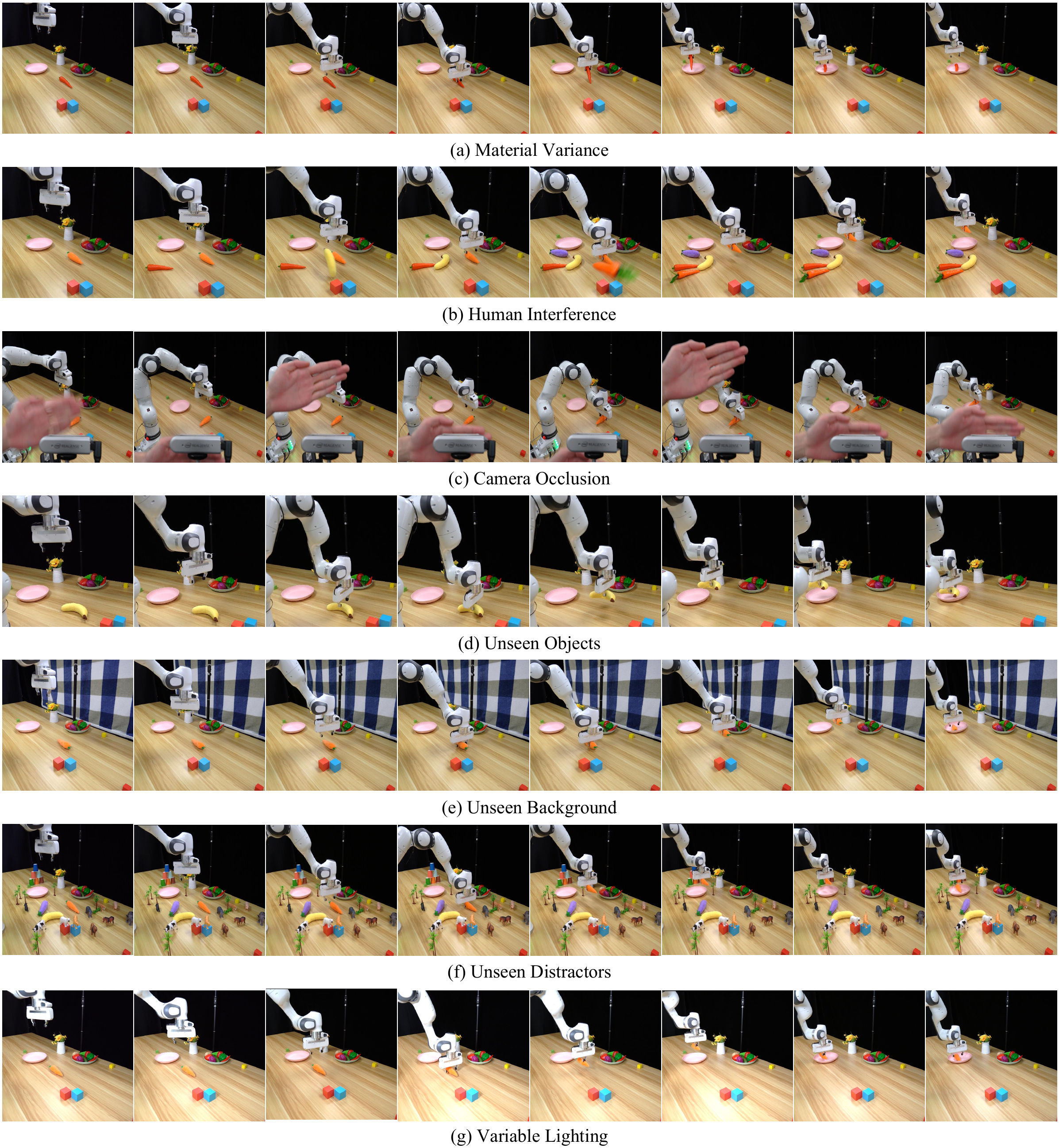}
    \caption{\textbf{Visualizations of real-world Generalization and Robustness tasks.}}
    \label{fig:visual_real_world_gene_robu}
\end{figure*}

\noindent\textbf{Generalization and Robustness.} We evaluate the generalization and robustness of different policies. All policies are evaluated on the \textit{Put the carrot on the plate} task and trained on the data collected in all seven tasks of \textit{Simple Pick-and-place}. Detailed success rates are shown in Tab.\ref{tab:generalization_results}, visualizations are shown in Fig.\ref{fig:visual_real_world_gene_robu} and our video demo:
\begin{itemize}
    \item \textbf{Material variance.} This experiment aims to evaluate the model’s generalization across objects. Carrots differing in appearance and material properties are used, as is shown in Fig.\ref{fig:visual_real_world_gene_robu} (a). We conduct the evaluation using the same spatial variance settings as in the \textit{Put the carrot on the plate} task.

    \item \textbf{Human interference.} This experiment assesses the model’s robustness under manual disturbances. In each test trial, irrelevant or similar objects are thrown to introduce interference, as shown in Fig.\ref{fig:visual_real_world_gene_robu} (b).

    \item \textbf{Camera occlusion.} This experiment evaluates the model’s robustness to incomplete observations. The camera view is occluded at a fixed frequency to simulate partial or corrupted input, as illustrated in Fig.\ref{fig:visual_real_world_gene_robu} (c).

    \item \textbf{Unseen objects.} This experiment assesses the model’s generalization to grasping different objects. We test objects not seen during training, placed in similar varying positions, as shown in Fig.\ref{fig:visual_real_world_gene_robu} (d).

    \item \textbf{Unseen background.} This experiment evaluates the model’s robustness to varying backgrounds. The background is changed from a plain white surface to a textured one, as shown in Fig.\ref{fig:visual_real_world_gene_robu} (e).
    
    \item \textbf{Unseen distractors.} This experiment assesses the impact of distractor objects on the model’s performance. A large number of unrelated additional objects, such as toys, are placed on the table, as shown in Fig.\ref{fig:visual_real_world_gene_robu} (f).

    \item \textbf{Variable lighting.} This experiment assesses the model’s robustness to varying lighting conditions by employing periodic flashing lights, as shown in Fig.\ref{fig:visual_real_world_gene_robu} (g).
\end{itemize}

\begin{table}[h]
\centering
\renewcommand{\arraystretch}{1.2}
\setlength{\tabcolsep}{4pt}
\vspace{2mm}
\resizebox{0.8\linewidth}{!}{
\begin{tabular}{l|ccccccc}
\toprule
\multirow{2}{*}{Method} 
& \multicolumn{7}{c}{\textbf{Generalization and Robustness}} \\
\cmidrule(lr){2-8}
& Material Var. & Human Interf. & Cam. Occlusion & Unseen Obj. & Unseen Bkg. & Distractors & Lighting \\
\midrule
DP3     & 72.0 & 24.0 & 12.0 & 52.0 & 64.0 & 28.0 & 60.0 \\
OpenVLA & 68.0 & 56.0 & 20.0 & 60.0 & 56.0 & 64.0 & 48.0 \\
\bottomrule \rowcolor{gray!15} 
Ours    & 80.0 & 76.0 & 64.0 & 72.0 & 72.0 & 76.0 & 68.0 \\
\bottomrule
\end{tabular}
}
\caption{\textbf{Performance comparison across real-world Generalization and Robustness tasks.}}
\label{tab:generalization_results}
\end{table}

\clearpage
\newpage

\section{Failure Cases}
\label{sec_Failure_Cases}

\begin{figure*}[h]
    \centering
    \includegraphics[width=1\linewidth]{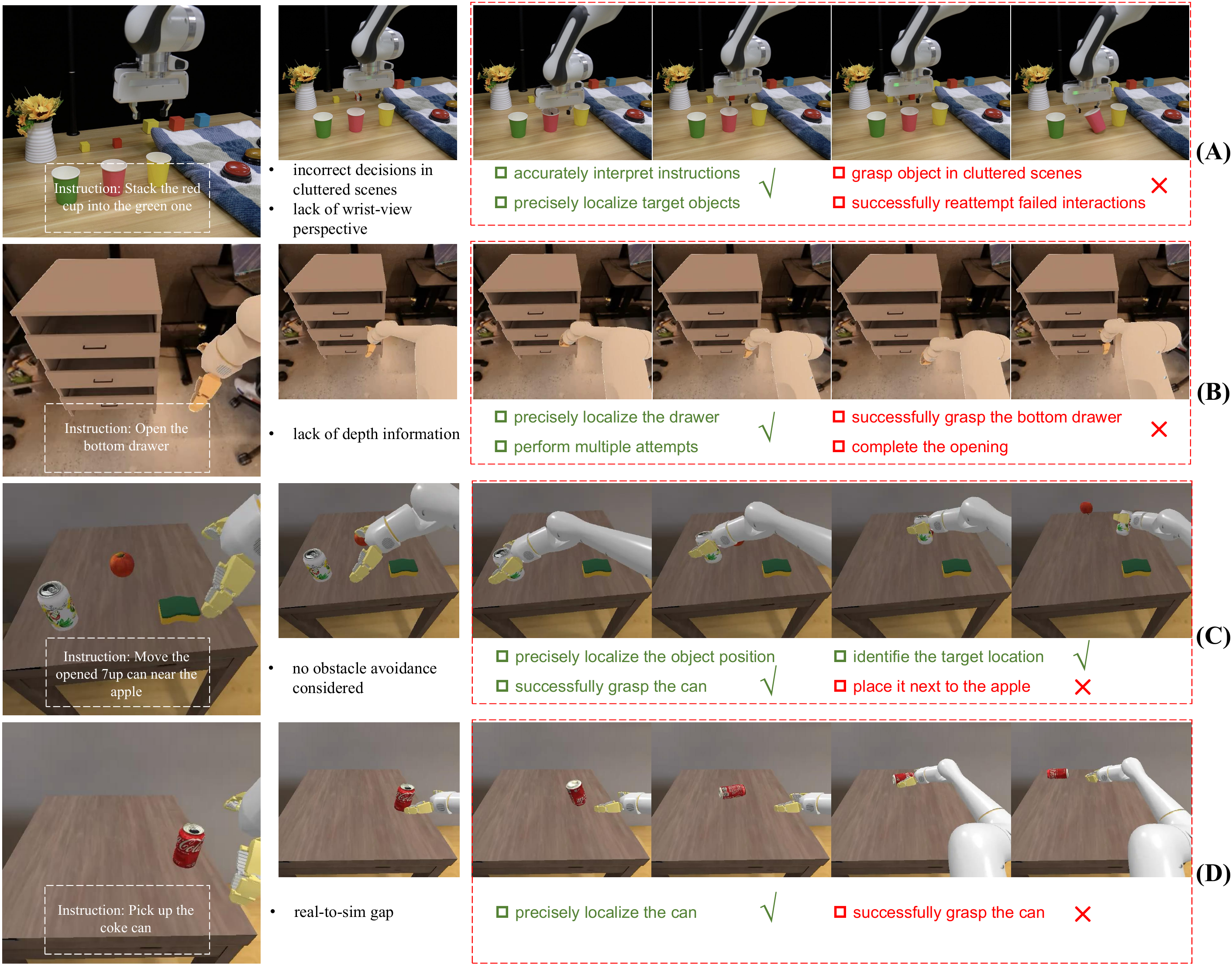}
    \caption{\textbf{Failure cases for CronusVLA} across real-world experiment and SimplerEnv benchmark.}
    \label{fig:case_study_failure_cases}
\end{figure*}
Despite the strong performance of CronusVLA in both simulation benchmarks and real-world experiments, notable failure cases remain, highlighting opportunities for further improvement. As illustrated in Fig.\ref{fig:case_study_failure_cases}, we present four representative examples. In (A), a real-world \textit{stack cups} task, the model reaches a graspable position but makes unnecessary adjustments. This may result from the limitations of a single third-person view, particularly in cluttered scenes where occlusions and viewpoint variance impair perception. Incorporating more cluttered-scene data and adding wrist-view input could improve robustness. In (B), the model struggles with depth estimation when opening distant drawers; despite multiple attempts, the model consistently fails to accurately grasp the bottom drawer. This underscores the importance of precise depth perception in such tasks. In (C), although CronusVLA selects the correct action, the robot arm knocks the target object (an apple) away due to occlusion-induced collisions. We attribute the failure to the model’s lack of explicit obstacle avoidance under occlusion. In (D), certain behaviors acceptable in the SimplerEnv simulation result in unrealistic physical outcomes, such as a slight touch causing a can to flip and roll excessively. This suggests a real-to-sim gap that needs further investigation.

\clearpage
\newpage

\section{Complexity Analysis} 
\label{sec_Complexity_Analysis}
For directly processing M past-moment frames, assuming the VLM splits each image into $\mathcal{P}$ tokens and the instruction into $\mathcal{I}$ tokens, the total number of tokens becomes $(M+1)\cdot \mathcal{P} + \mathcal{I}$, and the self-attention complexity of the language model increases from $O((\mathcal{P}+\mathcal{I})^2)$ to
\begin{equation}
    O(((M+1) \cdot \mathcal{P} + \mathcal{I})^2) \simeq O(M^2),
\end{equation}
where $\mathcal{P} \gg \mathcal{I}$, thus substantially increasing computational demands. Notably, since the action head or action detokenizer of former VLA models contains fewer parameters than the VLM backbone, the naive approach’s complexity is dominated by the quadratic term $O(M^2)$. 
In contrast, our approach processes each frame independently through the VLM and models multi-frame relationships at the feature level, only incurring a linear increase with the number of frames:
\begin{equation}
    O(M \cdot T_{VLM}) + O(M\cdot T_{decoder}) \simeq O(M \cdot T_{VLM}) \simeq O(M),
\end{equation}
where $T_{\text{VLM}}$ and $T_{\text{decoder}}$ denote the inference time of the VLM and the cross-frame decoder, respectively. As $T_{\text{decoder}} \ll T_{\text{VLM}}$, the overall complexity scales linearly with $M$, leading to limited impact on inference speed in practice.

\section{Limitation and Future Works}
\label{sec_Limitation_and_Future_Works}

\noindent\textbf{Efficient utilization of inter-frame information.} Inspired by the effectiveness of multi-frame inputs in low-level action prediction, we introduce a feature-level multi-frame modeling approach that utilizes learnable features extracted by the VLA module. However, in scenarios where observations and instructions remain largely unchanged across frames, large language models (LLMs) tend to process redundant tokens, leading to considerable spatial redundancy. Mitigating this redundancy by capturing inter-frame differences can enhance model efficiency and lower training costs. A promising direction involves exploring cache-like mechanisms that leverage inter-frame image differences and the compositional nature of LLMs to improve efficiency further.

\noindent\textbf{Language-driven embodied manipulation.} Most existing VLAs are fine-tuned from pretrained VLMs but often underexploit the explicit language reasoning capabilities that can guide manipulation, relying instead on implicit vision-language alignment. Similarly, CronusVLA does not explicitly leverage foundational language abilities during inference. In future work, we aim to better integrate language-driven reasoning with action generation to enhance manipulation performance.

\noindent\textbf{Toward a more powerful and generalizable architecture.} CronusVLA builds upon the original OpenVLA design, which processes only single-view images and language instructions, without incorporating proprioceptive states or multi-view observations. However, integrating richer modalities is essential for enhancing performance and generalization. In future work, we aim to extend the CronusVLA framework to support multi-view and state-conditioned temporal modeling for broader applicability across diverse embodied tasks.

\section{Broader Impact}
\label{sec_Broader_Impacts}
\noindent\textbf{Social Impact.} This paper aims to enhance the temporal understanding of vision-language-action (VLA) models for long-horizon robotic manipulation tasks. Improved temporal modeling can benefit assistive robotics in daily life and industrial settings by enabling robots to interpret and execute complex, time-extended instructions more effectively.

\clearpage
\newpage

\section{More Backgrounds}
\label{sec_More_Backgrounds}

\textbf{Vision-language data.} Our 7B model is based on the Prismatic~\cite{prismatic} backbone, which combines the LLaMA2~\cite{llama2} language model with DINOv2~\cite{dinov2} and SigLIP~\cite{siglip} as vision encoders. LLaMA2 is pretrained on approximately 2 trillion language tokens, while DINOv2 and SigLIP are trained on large-scale visual datasets. Prismatic is trained for vision-language alignment using datasets such as LAION~\cite{schuhmann2021laion}, Conceptual Captions~\cite{sharma2018conceptual}, and SBU Captions~\cite{ordonez2011im2text}, totaling roughly 558K image-caption pairs. It is further instruction-tuned on 665K multimodal data, including LLaVA Synthetic Data~\cite{llava}, VQA datasets~\cite{goyal2017making,hudson2019gqa}, referring expression datasets~\cite{kazemzadeh2014referitgame,krishna2017visual}, and so on.

\textbf{Vision-language-action data.} For VLA data, we follow the pretraining protocols of Octo~\cite{octo} and OpenVLA~\cite{openvla}, utilizing 27 datasets from Open X-Embodiment~\cite{oxe}. These datasets primarily comprise single-arm demonstrations with at least one third-person observation and corresponding end-effector actions. Note that some subsets, such as Kuka~\cite{kalashnikov2018scalable}, may contain imperfect language instructions. And as shown in~\cite{cogact,openvla}, Droid~\cite{droid} and Language Table~\cite{lynch2023interactive} datasets may lead to a potential out-of-distribution issue. Especially, the Fractal dataset is a large-scale collection of open-world manipulation demonstrations, comprising approximately 87k episodes and 3.8M images. These demonstrations were collected using the Google Robot platform, providing a diverse set of real-world robotic interactions. Bridge-v2 is another extensive dataset designed to facilitate scalable robot learning. It contains 60k trajectories and 2.1M images collected across different environments using the WidowX Robot platform. All data is stored in the RLDS format~\cite{ramos2021rlds} and mixed using predefined ratios during training.

\textbf{Training for discrete action.} The cross-entropy loss function is defined as: $\mathcal{L}_{\text{CE}} = -\sum_{t=1}^{T} \log P_{\theta}(y_t \mid y_{<t}, x)$, where $y_t$ represents the target token at position $t$, $y_{<t}$ denotes the sequence of preceding tokens, $x$ encapsulates the input modalities (including images and language instructions), and $P_{\theta}$ is the model’s predicted probability distribution parameterized by $\theta$. For action prediction, we follow RT-2~\cite{RT2} and OpenVLA~\cite{openvla}, employing a discretization strategy wherein each dimension of the robot’s continuous action space is divided into 256 uniform bins. This transforms continuous action values into discrete tokens, facilitating their integration into the language modeling framework. To accommodate these action tokens within the model’s vocabulary, OpenVLA repurposes the 256 least frequently used tokens in the Llama tokenizer’s vocabulary, replacing them with the newly introduced action tokens.

\textbf{SimplerEnv.} 
In the simulation, we conduct the majority of our experiments within SimplerEnv \cite{simplerenv}, a benchmark designed to evaluate the capabilities of models in performing various tasks with the WidowX Robot and the Google Robot environment, which can effectively evaluation the cross-embodiment generalization and exhibit a strong performance correlation between simulator and real-world scenarios. For the Google Robot environment, SimplerEnv includes two experimental settings. Visual Matching (VM) strictly follows a real-to-sim replication, assessing the model’s ability to learn demonstration trajectories while testing its spatial generalization through viewpoint and position variations. Variant Aggregation (VA) introduces environmental variations in background, lighting, distractors, and textures, evaluating the policy’s adaptability and robustness in a zero-shot manner. Both of them primarily include tasks such as picking a coke can, moving near, opening/closing the drawer, opening the drawer, and placing the apple in. For the WidowX Robot environment, only the Visual Matching (VM) setting is used, ensuring a faster evaluation aligned with Bridge-v2~\cite{bridgev2}. It mainly includes tasks such as putting the carrot on the plate, putting the spoon on the towel, stacking the green block on the yellow block, and putting the eggplant in the yellow basket. We evaluate the SimplerEnv following the setting in ~\cite{cogact}.

\textbf{LIBERO.} It comprises four task suites, including LIBERO-Spatial, LIBERO-Object, LIBERO-Goal, and LIBERO-Long, to assess specific aspects of knowledge transfer in lifelong robot learning. For LIBERO-Spatial, it focuses on spatial reasoning, involving manipulating identical objects where the primary challenge lies in discerning spatial relationships. LIBERO-Object emphasis is on the object-centric manipulation, requiring the robot to interact with unique objects and highlighting the transfer of object-specific knowledge. LIBERO-Goal evaluates goal-conditioned adaptability, requiring the robot to adjust its actions to achieve different goals, testing its understanding of task objectives. LIBERO-Long is designed to challenge long-term planning and execution, comprising multi-stage tasks that require sustained attention and coordination. Each task suite comprises 10 tasks with 50 human-teleoperated demonstrations. We follow the data cleaning procedure of ~\cite{openvla}, which filters out corrupted samples and standardizes the data format to RLDS.

\clearpage
\newpage

\end{document}